\documentclass{article}

\usepackage{arxiv}
\usepackage{breakcites}
\usepackage{cite}

\usepackage{times}

\usepackage[utf8]{inputenc} %
\usepackage[T1]{fontenc}    %
\usepackage{hyperref}       %
\usepackage{url}            %
\usepackage{booktabs}       %
\usepackage{amsfonts}       %
\usepackage{nicefrac}       %
\usepackage{microtype}      %
\usepackage{algorithm}
\usepackage{algpseudocode}
\usepackage{bm}
\usepackage{amssymb}
\usepackage{amsmath,empheq, amsfonts}
\usepackage{multicol}
\usepackage{multirow}
\usepackage{booktabs}
\usepackage{wrapfig}
\usepackage{makecell}
\usepackage{graphicx}
\usepackage{tikz}
\usepackage{colortbl}
\usepackage{subcaption}
\usepackage{placeins}
\usepackage{tabu}
\usepackage{microtype}

\DeclareMathOperator{\Hess}{Hess}
\DeclareMathOperator{\Tr}{Tr}

\DeclareMathOperator{\diag}{diag}
\DeclareMathOperator{\range}{range}

\DeclareMathOperator{\tr}{\mathrm{Tr}}
\DeclareMathOperator{\KL}{\mathrm{KL}}

\DeclareMathOperator{\St}{St}

\newcommand{\tU}{\tilde{U}}

\newcommand{\tabfigure}[2]{\raisebox{-.5\height}{\includegraphics[#1]{#2}}}

\newtheorem{lemma}{Lemma}

\newtheorem{proposition}{Proposition}

\title{Disentangled Representation Learning and Generation with Manifold Optimization}

\author{%
 Arun Pandey\\
  Department of Electrical Engineering\\
  ESAT-STADIUS, KU Leuven\\
  Kasteelpark Arenberg 10, B-3001 Leuven, Belgium\\
  \texttt{arun.pandey@esat.kuleuven.be} \\
   \And
 Micha\"el Fanuel\\
  Université de Lille, CNRS, Centrale Lille,\\
UMR 9189 – CRIStAL, F-59000 Lille, France.\\
  \texttt{michael.fanuel@univ-lille.fr}
   \And
 Joachim Schreurs\\
  Department of Electrical Engineering\\
  ESAT-STADIUS, KU Leuven\\
  Kasteelpark Arenberg 10, B-3001 Leuven, Belgium\\
  \texttt{joachim.schreurs@esat.kuleuven.be}
   \And
Johan A. K. Suykens\\
  Department of Electrical Engineering\\
  ESAT-STADIUS, KU Leuven\\
  Kasteelpark Arenberg 10, B-3001 Leuven, Belgium\\
  \texttt{johan.suykens@esat.kuleuven.be}
}

\begin{document}
\maketitle
%
%Abstract
\begin{abstract}
Disentanglement is a useful property in representation learning which increases the interpretability of generative models
such as Variational autoencoders (VAE), Generative Adversarial Models, and their many variants. Typically in such models, an increase in disentanglement performance is traded-off with generation quality.
In the context of latent space models, this work presents a representation learning framework that explicitly promotes disentanglement by encouraging orthogonal directions of variations.
The proposed objective is the sum of an autoencoder error term along with a Principal Component Analysis reconstruction error in the feature space. This has an interpretation of a Restricted Kernel Machine {with the eigenvector matrix valued on} the Stiefel manifold.
Our analysis shows that such a construction promotes disentanglement by matching the principal directions in the latent space with the directions of orthogonal variation in data space. In an alternating minimization scheme, we use Cayley ADAM algorithm --- a stochastic optimization method on the Stiefel manifold along with the ADAM optimizer.
Our theoretical discussion and various experiments show that the proposed model improves over many VAE variants in terms of both generation quality and disentangled representation learning.
\end{abstract}
%%%%%%%%%%%

\section{Introduction~\label{sec:introduction}}
Latent space models are popular tools for sampling from high-dimensional distributions. Often, only a small number of latent factors are sufficient to describe data variations. These models exploit the underlying structure of the data and learn explicit representations that are faithful to the data generating factors.  Popular latent space models are Variational autoencoders (VAEs)~\cite{kingma_auto_encoding_2013}, Restricted Boltzmann Machines (RBMs)~\cite{salakhutdinov_deep}, Normalizing Flows~\cite{pmlr-v37-rezende15}, and their many variants.

\begin{figure}[htbp]
	\centering
	\def\svgwidth{0.9\linewidth}
	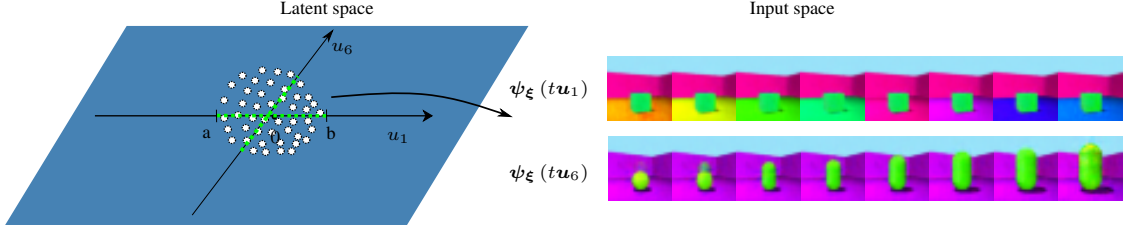
	\caption{Images by the decoder of the latent space traversal, i.e., $\bm{\psi}_{\bm{\xi}}\left(t \bm{u}_i\right)$ for $t\in [a,b]$ with $a<b$ and for some $i\in\{1,\dots, m\}$. Green and black dashed lines represent the walk along $\bm{u}_1$ and $\bm{u}_6$ respectively. At every step of the walk, {the output of the decoder generates} the data in the input space. The images were generated by $\St$-RKM with $\sigma = 10^{-3}$ on 3Dshapes dataset; see Figure~\ref{fig:traversal_imgs} for traversal along other components.}
	\label{fig:traversal_rkm}
\end{figure}

In latent variable models, one is often interested in modeling the data in terms of \emph{uncorrelated} or \emph{independent} components, yielding a so-called `disentangled' representation~\cite{bengio2013representation} which is often studied in the context of VAEs.
Generative Adversarial Networks (GAN) have also been extended to perform disentangled representation learning, for instance, with Info-GANs.
It is a GAN that also maximizes the mutual information between a small subset of the discrete latent codes and the true images.
In principle, disentanglement corresponds to {identifying the underlying factors which generate the data}.
Components corresponding to the orthogonal directions in latent space may be interpreted as generating distinct factors in the input space e.g. lighting conditions, style, colors, etc.
An illustration of a latent traversal is shown in Figure~\ref{fig:traversal_rkm}, where one observes that only one specific feature of the image is changing as one moves along a component in the latent space. For instance, in Figure~\ref{fig:traversal_rkm}, we observe that moving along the first {component (vector $ \bm{u}_{1}  $)} generates images where only floor color is varying while all other features such as shape, scale, wall color, object color, etc. are constant.
	{Whereas traversing along the sixth component (vector $ \bm{u}_{6} $) for instance, generates images where only the object scale changes as shown in the second row.
		As we explain later, the components here refer to the principal components given by the Principal Component Analysis (PCA). Therefore these principal directions encode the directions of maximum variance. Since the floor color is encoded by the largest number of pixels, it gets represented by the first principal component $ \bm{u}_{1}  $. Similarly, the other components correspond to the directions with smaller variance.}
An advantage of such a representation is that the different latent units impart more interpretability to the model. Disentangled models are useful for the generation of plausible pseudo-data with certain desirable properties, e.g. generating  new car designs with a predefined color or height.

Now we introduce the mathematical setting to formalize our discussion throughout the paper. {We start by introducing a VAE~\cite{kingma_auto_encoding_2013}}.
Let $p(\bm{x})$ be the distribution of the data $\bm{x}\in\mathbb{R}^d$ and consider latent vectors $\bm{z}\in \mathbb{R}^\ell$ with the prior distribution $p(\bm{z})$, typically a standard normal distribution.
Then, one defines an encoder $q(\bm{z}|\bm{x})$ that can be {deterministic or probabilistic}, for e.g. given by $ \mathcal{N}(\bm{z}|\bm{\phi_{\theta}}(\bm{x}),\gamma^2\mathbb{I})$, where the mean\footnote{{A typical implementation of VAE includes another neural network (after the primary network) for parametrizing the covariance matrix.} To simplify this introductory discussion, this matrix is here chosen as a constant diagonal $\gamma^2\mathbb{I}$.} is given by the neural network $\bm{\phi_{\theta}}$ parametrized by $\bm{\theta}$.
A random decoder $p(\bm{x}|\bm{z}) =  \mathcal{N}(\bm{x}|\bm{\psi_{\xi}}(\bm{z}),\sigma^2_0\mathbb{I})$ is associated to the decoder neural network $\bm{\psi_{\xi}}$, parametrized by $\bm{\xi}$, which maps latent codes to the data points.
A VAE is {trained by maximizing the lower bound to the idealized log-likelihood as below}:
\begin{equation}
	\mathbb{E}_{\bm{z}\sim q(\bm{z}|\bm{x})}[\log (p(\bm{x}|\bm{z}))]- \beta  \KL(q(\bm{z}|\bm{x}),p(\bm{z})) \leq \log p(\bm{x}).  \label{eq:ELBO_VAE}
\end{equation}
{This lower bound is often called as} the Evidence Lower Bound (ELBO) when $\beta = 1$. \cite{higgins2017beta} show that the larger values of $\beta>1$ promote more disentanglement but at the expense of generation quality. In this paper, we attempt to reconcile the generation quality with disentanglement.
To introduce the model, we first make explicit the connection between $\beta$-VAEs and standard autoencoders (AEs).
Let the dataset  be $\{ \bm{x}_{i}\}_{i=1}^{n} \text{~with~}\bm{x}_{i} \in \mathbb{R}^d $.
Let $q(\bm{z}|\bm{x}) = \mathcal{N}(\bm{z}|\bm{\phi_{\theta}}(\bm{x}),\gamma^2\mathbb{I})$ be an encoder, where $\bm{z}\in \mathbb{R}^\ell$. For a fixed $\gamma>0$,  the maximization problem \eqref{eq:ELBO_VAE} is then equivalent to the minimization of the regularized AE
\begin{equation}
	\min_{\bm{\theta}, \bm{\xi}}\frac{1}{n}
	\sum_{i=1}^{n}\Big\{ \mathbb{E}_{\bm{\epsilon}}\|\bm{x}_i - \bm{\psi_{\xi}}(\bm{\phi_{\theta}}(\bm{x}_i)+\bm{\epsilon})\|_2^2  +\alpha\|\bm{\phi_{\theta}}(\bm{x}_i)\|_2^2\Big\} ,\label{eq:VAE_AE}
\end{equation}
where $\alpha = \beta\sigma^2_0$, $\bm{\epsilon}\sim\mathcal{N}(0,\gamma^2\mathbb{I})$ and where additive constants depending on $\gamma$ have been omitted. The first term in~\eqref{eq:VAE_AE} can be interpreted as an AE loss whereas the second term can be viewed as a regularization. This regularized AE interpretation motivates our method as introduced in Section \ref{sec:Proposed}.

The rest of the paper is organized as follows. In Section~\ref{sec:related_works} we discuss the closely related work on disentangled representation learning and generation in the context of autoencoders.
Further in Section~\ref{sec:Proposed} we describe the proposed model along with the connection between PCA and disentanglement.
In Section~\ref{sec:contributions} we discuss our contributions.
In Section~\ref{sec:elbo},  we derive the evidence lower bound of the proposed model and show  connections with the probabilistic models.
In Section~\ref{sec:exp}, we describe our experiments and discuss the results.

\begin{figure}[ht]
	\centering
	\def\svgwidth{0.9\linewidth}
	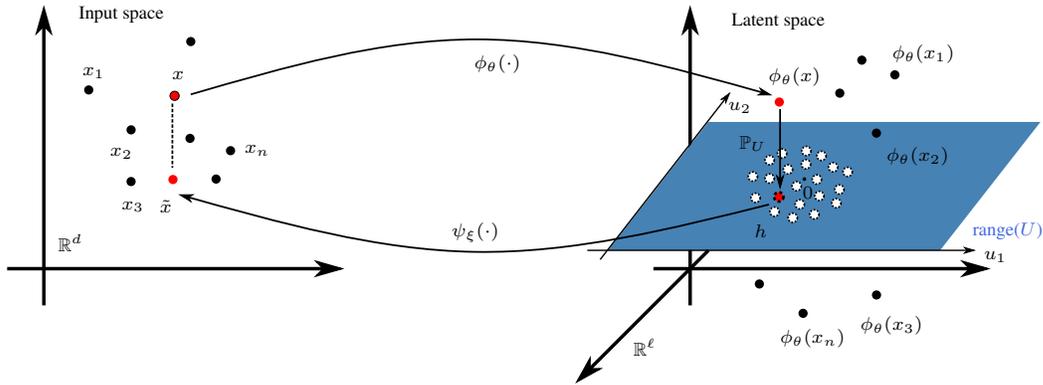
	\caption{Schematic illustration of $\St$-RKM training problem. The length of the dashed line represents the reconstruction  error (see autoencoder term in \eqref{eq:ReducedObjective}) and the length of the vector projecting on hyperplane represents the PCA reconstruction error. After  training, the projected points tend to be distributed normally on the hyperplane. }
	\label{fig:schematic_diagram}
\end{figure}
\section{Related work \label{sec:related_works}}
Related works can be broadly classified into two categories: Variational autoencoders (VAE) in the context of disentanglement and Restricted Kernel Machines (RKM), a recently proposed modeling framework that integrates kernel methods with deep learning.

\noindent  \textbf{VAE}: As discussed in the introduction, \cite{higgins2017beta} suggested that a stronger emphasis on the posterior to match the factorized unit Gaussian prior puts further constraints on the implicit capacity of the latent bottleneck. \cite{burgess2018understanding} further analyzed the effect of the $\beta$ term in depth. Later \cite{MIG_VAE} showed that the KL term includes the Mutual Information Gap which encourages disentanglement.
Recently, several variants of VAEs promoting disentanglement have been proposed by adding extra terms to the ELBO. For instance, FactorVAE~\cite{FactorVAE} augments the ELBO by a new term enforcing factorization of the  marginal posterior (or aggregate posterior).
\cite{Robinek} analyzed the reason for the alignment of the latent space with the coordinate axes, as the design of VAE itself does not suggest any such mechanism. The authors argue that due to the diagonal approximation in the encoder together with the inherent stochasticity forces the local orthogonality of the decoder. \cite{Locatello2020Disentangling} considered adding an extra term that accounts for the knowledge of some partial label information to improve disentanglement.
Later \cite{ghosh2019variational} studied the deterministic AEs, where another quadratic regularization on the latent vectors was proposed. In contrast to~\cite{Robinek} where the implicit orthogonality of VAE was studied, our proposed model has orthogonality by design due to the introduction of the Stiefel manifold.\\

\noindent\textbf{RKM\label{sec:RKM}}:
Restricted Kernel Machines (RKM)~\cite{suykensdeep2017} provides a representation of kernel methods with visible and hidden variables similar to the energy function of Restricted Boltzmann Machines (RBM)~\cite{lecun_learning_2004, hinton2005}, thus linking kernel methods with RBMs.
Training and prediction schemes are characterized by the stationary points for the unknowns in the objective.
The equations in these stationary points lead to solving a linear-system or matrix decomposition for the training. \cite{suykensdeep2017} shows various RKM formulations for doing classification, regression, kernel PCA and Singular Value Decomposition.
Later the Kernel PCA formulation of RKM was extended  to a multi-view generative model called Generative-RKM  (Gen-RKM) which  uses Convolutional Neural Networks as explicit feature maps~\cite{GENRKM,robust2020}. For the joint feature selection and subspace learning,
the proposed training procedure performs eigendecomposition of the kernel/covariance matrix in every mini-batch of the optimization scheme. Intuitively, the model could be seen as learning an autoencoder with Kernel PCA in the bottleneck part.
As a result, the computational complexity scales cubically with the mini-batch size and is proportional to the number of mini-batches.
Moreover, backpropagation through the eigendecomposition could be  numerically unstable due to the possibility of small eigenvalues. All such limitations are addressed by our proposed model.

\section{Proposed mechanism\label{sec:Proposed}}

The main idea of this paper consists of learning an autoencoder along with finding an optimal linear subspace of the latent space such that the variance of the training set in latent space is maximized within this space.
See Figure \ref{fig:schematic_diagram} to follow the discussion below.
	{Note the distinction with linear autoencoders which also project the data into the low-dimensional subspace although via non-orthogonal transformations.
		As a consequence, the latent variables are not guaranteed to be uncorrelated.}
The encoder $\bm{\phi_{\theta}}:\mathbb{R}^d \to \mathbb{R}^\ell$ typically sends input data to a latent space while the decoder $\bm{\psi_{\xi}}: \mathbb{R}^\ell \to \mathbb{R}^d$ goes in the reverse direction, and constitutes an approximate inverse.
Both the encoder and decoder are neural networks parameterized by vectors $\bm{\theta}$ and $\bm{\xi}$.
However, it is unclear how to define a parametrization or an architecture of these neural networks so that the learned representation is disentangled.
Therefore, in addition to these trained parameters, we also jointly find an $m$-dimensional linear subspace $\range(U)$ of the latent space $\mathbb{R}^\ell$, such that the encoded training points mostly lie within this subspace.
This linear subspace is given by the span of the orthonormal columns of the $\ell\times m$ matrix $U = [\bm{u}_1, \dots ,\bm{u}_m]$.
The set of such matrices with $m$ orthonormal columns in $\mathbb{R}^\ell$ with $\ell\geq m$ defines the Stiefel manifold $\St(\ell,m)$. For a reference about optimization on Stiefel manifold,  we refer to~\cite{AbsilBook}.
Input data is then encoded into a subspace of the latent space by
\begin{equation*}
	\bm{x}\mapsto \mathbb{P}_{U}\bm{\phi_{\theta}}(x)= \bm{u}_1^\top \bm{\phi_{\theta}}(x)\times
	\begin{bmatrix}
		|        \\
		\bm{u}_1 \\
		|
	\end{bmatrix}+ \dots + \bm{u}_m^\top \bm{\phi_{\theta}}(x) \times\begin{bmatrix}
		|        \\
		\bm{u}_m \\
		|
	\end{bmatrix}  ,
\end{equation*}
where the {orthogonal} projector onto $\range(U)$ is simply  $\mathbb{P}_{U} = U U^\top$.
\paragraph{Orthogonal latent directions:} Naturally, given an $m \times m$ orthogonal matrix $O$ and a matrix $U\in \St(\ell,m)$, we have \[\range(U) = \range(UO).\]
To select a specific matrix $U_\star = [\bm{u}_{\star,1}, \dots, \bm{u}_{\star,m}]\in \St(\ell,m)$, we choose $\bm{u}_{\star,1}, \dots ,\bm{u}_{\star,m}$ to be the eigenvectors of the matrix
$ C_{\bm{\theta}} = \frac{1}{n}\sum_{i=1}^n  \bm{\phi}_{\bm{\theta}}(\bm{x}_i)\bm{\phi}_{\bm{\theta}}^\top(\bm{x}_i),$
associated with the $m$ largest eigenvalues sorted in descending order.
For simplicity, we assume that  the $m$ largest eigenvalues of $C_{\bm{\theta}}$ are distinct, whereas the general case involves minor technicalities.
Here the feature map is assumed to be centered, i.e. $\mathbb{E}_{\bm{x}\sim p(\bm{x})} [\bm{\phi}_{\bm{\theta}}(\bm{x})] = \bm{0}$, so that $C_{\bm{\theta}}$ is interpreted as a covariance matrix.
Next, we state a result that will be used extensively later.
\begin{proposition}\label{Prop:U}
	Let $M$ be an $\ell \times \ell$ symmetric matrix. Let $\nu_1, \dots, \nu_m$ be its $m$ smallest eigenvalues, possibly including multiplicities, with associated orthonormal eigenvectors $\bm{v}_1, \dots, \bm{v}_m$. Let $V$ be a matrix whose columns are these eigenvectors. Then, the optimization problem
	$\min_{U\in\St(\ell,m)} \tr(U^\top M U)$ has a minimizer at  $U_\star = V$ and we have
	${U_\star}^\top M U_\star = \diag(\bm{\nu}),$
	with $\bm{\nu} = (\nu_1, \dots, \nu_m)^\top$.
\end{proposition}
A few remarks are as follows. First, if $U_\star$ is a minimizer of the optimization problem in Proposition~\ref{Prop:U} then ${U'}_\star = U_\star O$ with $O$ orthogonal is also a minimizer, but ${U'}_\star^\top M {U'}_\star$
is not necessarily diagonal. Second, notice that, if the eigenvalues of $M$ in Proposition~\ref{Prop:U} have a multiplicity larger than $1$, there can exist several sets of eigenvectors $\bm{v}_1, \dots, \bm{v}_m$, associated to the $m$ smallest eigenvalues, spanning distinct linear subspaces.
Nevertheless, in practice, the eigenvalues of the matrices considered in this paper are numerically distinct.

Let us now use Proposition~\ref{Prop:U}. For a given positive integer $m\leq \ell$, the subspace spanned by the eigenvectors of $C_{\bm{\theta}}$ with the $m$ largest eigenvalues is obtained by solving
\begin{equation*}
	\min_{U\in\St(\ell,m)} \Tr\left(C_{\bm{\theta}} -\mathbb{P}_U C_{\bm{\theta}} \mathbb{P}_U \right)= \frac{1}{n}\sum_{i=1}^n \| \mathbb{P}_{U^\perp} \bm{\phi}_{\bm{\theta}}(\bm{x}_i)\|_2^2,
\end{equation*}
where $\mathbb{P}_{U^\perp} = \mathbb{I}-\mathbb{P}_{U}$,
as it is explained, for instance, in Section~4.1 of~\cite{Woodruff}. The above objective
corresponds to the reconstruction error of Kernel PCA, for the kernel $k_{\bm{\theta}}(\bm{x},\bm{y}) =  \bm{\phi}^\top_{\bm{\theta}}(\bm{x}) \bm{\phi}_{\bm{\theta}}(\bm{y}) $.
As described earlier, we choose a specific $U_\star\in \St(\ell,m)$ by requiring that the following matrix is diagonal
\begin{equation}
	U^\top_\star C_{\bm{\theta}} U_\star = \diag(\bm{\lambda}),
	\label{eq:svd}
\end{equation}
where $\bm{\lambda}$ is a vector containing the $m$ largest eigenvalues sorted in decreasing order. If these eigenvalues are distinct, then the $U_\star$ is essentially unique, up to sign flip of each of its columns. Notice that $\Tr(U^\top_\star C_{\bm{\theta}} U_\star) = \Tr(U_\star U^\top_\star C_{\bm{\theta}} U_\star U^\top_\star)$.
\paragraph{Orthogonal directions of variation in input space:} We want the lines defined by the orthonormal vectors $\{\bm{u}_{\star,1}, \dots, \bm{u}_{\star,m}\}$ to provide directions associated with different generative factors of our  model.
In other words, we conjecture that a possible formalization of disentanglement is that   \emph{the principal directions in latent space match orthogonal directions of variation in the data space} (see Figure~\ref{fig:schematic_diagram}).
That is to say, we would like that
\begin{equation}
	U^\top_\star \sum_{a=1}^d\left(\nabla\bm{\psi}_a(\bm{y}_i)\nabla\bm{\psi}_a(\bm{y}_i)^\top\right) U_\star \text{~is diagonal},\label{eq:UNabla}
\end{equation}
for all the points in latent space $\bm{y}_i = \mathbb{P}_{U}\bm{\phi_{\theta}}(\bm{x}_i)$ for $i=1,\dots, n$. In~\eqref{eq:UNabla}, $\bm{\psi}_a(\bm{y})$ refers to the $a$-th component of the image $\bm{\psi}(\bm{y})\in \mathbb{R}^d$.
To sketch this idea, we study below the local motions in the latent space.

Let ${\Delta}_k =  \nabla \bm{\psi}(\bm{y})^\top\bm{u}_{\star,k} \in \mathbb{R}^d$ be the directional derivative of $\bm{\psi}$ at point $\bm{y}$ in the direction $\bm{u}_{\star,k}$ with $1\leq k\leq m$.
Then, as one moves in the latent space from {a point} $\bm y$ in the direction of $\bm{u}_{\star,k}$, the generated data changes by
\[\bm{\psi}(\bm{y}+ t \bm{u}_{\star,k}) - \bm{\psi}(\bm{y}) = t {\Delta}_k + \mathcal{O}(t^2),\]
with ${\Delta}_k \in \mathbb{R}^d$ and $t\in \mathbb{R}$. {Consider now a different direction, i.e., $k'\neq k$.
		As the latent point moves along $\bm{u}_{\star, k}$ or along $\bm{u}_{\star,k'}$, we expect the decoder output to vary in a significantly different manner, i.e., ${\Delta}_k^\top{\Delta}_{k'} = 0$.}
	{We presume this interpretation to model the change in floor color and object scale in Figure~\ref{fig:traversal_rkm} for instance.
		More explicitly, we can expect $\bm{u}_k$ and $\bm{u}_{k'}$ to model, respectively, the change of colors of the floor and of the main object while leaving the color of the other objects unchanged.
		Since the floor and the main object do not overlap, that is, are different regions in pixel space, we would have ${\Delta}_k^\top{\Delta}_{k'} = 0$.
		Admittedly, the change in object shape in Figure~\ref{fig:traversal_rkm} is less obviously  interpreted.}
Now, denote by ${\Delta}$ the matrix obtained by stacking the vector ${\Delta}_k$ as columns for $1\leq k \leq m$.
Explicitly, we have $\Delta = \nabla\bm{\psi}_a(\bm{y})^\top U_\star$.
\emph{Hence, for all $\bm y$ in the latent space, we expect the Gram matrix ${\Delta}^\top{\Delta}$ to be diagonal} (cf.~\eqref{eq:UNabla}).
We now discuss how this idea might be realized by minimizing specific objective functions.
\subsection{Objective function\label{sec:obj}} In this paper, we propose to train an objective function which is composed of (i) an AE loss
and (ii) a PCA loss.
Hence, the proposed model is given by
\begin{equation}
	\min_{\substack{ U\in \St(\ell,m)\\\bm{\theta}, \bm{\xi}}}\lambda\underbrace{\frac{1}{n}\sum_{i=1}^{n} L_{\bm{\xi},\mathbb{P}_{U}}\left(\bm{x}_i,\bm{\phi}_{\bm{\theta}}(\bm{x}_i)\right)}_{\text{Autoencoder objective}} + \underbrace{\Tr\left(C_{\bm{\theta}} -\mathbb{P}_U C_{\bm{\theta}} \mathbb{P}_U \right)}_{\text{PCA objective}},\label{eq:ReducedObjective}
\end{equation}
where $\lambda>0$ is a trade-off parameter
and $
	C_{\bm{\theta}} = \frac{1}{n}\sum_{i=1}^n  \bm{\phi}_{\bm{\theta}}(\bm{x}_i)\bm{\phi}_{\bm{\theta}}^\top(\bm{x}_i)$.
Naturally, the above objective is invariant if $U$ is replaced by $UO$ with $O$ an orthogonal matrix.
Given a local minimizer, we select $U_\star\in \St(\ell,m) $ such that 	$U^\top_\star C_{\bm{\theta}} U_\star$ is diagonal as in equation~\eqref{eq:svd} above, to identify the principal directions in the latent space.
This last step is conveniently done with a singular value decomposition; see  step $10$ of Algorithm~\ref{algo}. In the proposed model, reconstruction of an out-of-sample point $\bm{x}$ is given by $ \bm{\psi}_{\bm{\xi}}\big(\mathbb{P}_U\bm{\phi}_{\bm{\theta}}(\bm{x}) \big)$.
We call the procedure to
\begin{equation}
	\text{find a triplet  } (U_\star, \bm{\theta}, \bm{\xi}) \text{ solving \eqref{eq:ReducedObjective} s.t. } U^\top_\star C_{\bm{\theta}} U_\star \text{ is diagonal},\tag{St-RKM}
\end{equation}
the training of a Stiefel-Restricted Kernel Machines~\eqref{eq:ReducedObjective} in view of our discussion in Section~\ref{sec:RKM}.
The basic idea is to design different AE losses with a regularization term that penalizes the feature map in the orthogonal subspace $U^\perp$.
The choice of the AE losses is motivated by the expression of the regularized AE in~\eqref{eq:VAE_AE} and by the following Lemma which extends the result of~\cite{Robinek}. Here we adapt it in the context of optimization on the Stiefel manifold; see Appendix for the proof.
\begin{lemma}\label{Lemma:Smoothness}
	Let $\bm{\epsilon}\sim\mathcal{N}(\bm{0},\mathbb{I}_m)$ a random vector and $U\in\St(\ell,m)$. Let $\bm{\psi}_a(\cdot)\in \mathcal{C}^2(\mathbb{R^\ell})$ with $a\in [d]$. If the function $[\bm{\psi}(\cdot)-\bm{x}]^2_a$ has {$L_a$}-Lipschitz continuous Hessian for all $a\in [d]$, we have
	\begin{align}
		\label{eq:Smoothness}
		\mathbb{E}_{\bm{\epsilon}} \|\bm{x} - \bm{\psi}(\bm{y}+\sigma  U\bm{\epsilon})\|_2^2
		 & = \|\bm{x} - \bm{\psi}(\bm{y})\|_2^2  +\sigma^2  \Tr\big(U^\top\nabla\bm{\psi}(\bm{y})\nabla\bm{\psi}(\bm{y})^\top U\big)                   \\
		 & -\sigma^2 \sum_{a=1}^d[ \bm{x}- \bm{\psi}(\bm{y})]_a \Tr\big(U^\top\Hess_{\bm{y}} [\bm{\psi}_a] U \big)+ \sum_{a=1}^d R_a(\sigma),\nonumber
	\end{align}
	with $|R_a(\sigma)|\leq \frac{1}{6}\sigma^3 {L_a} \frac{\sqrt{2}(m+1) \Gamma((m+1)/2)}{\Gamma(m/2)}$ where $\Gamma$ is Euler's Gamma function.
\end{lemma}

A few remarks are as follows.
In Lemma \ref{Lemma:Smoothness}, the first term on the right-hand side in~\eqref{eq:Smoothness} plays the role of the classical AE loss. The second  term is proportional to the trace of~\eqref{eq:UNabla}. This is related to our discussion above where we argue that jointly  diagonalizing both $U^\top\nabla\bm{\psi}(\bm{y})\nabla\bm{\psi}(\bm{y})^\top U$ and $U^\top C_{\bm{\theta}} U$  helps to enforce disentanglement.
However, determining the behavior of the third term in~\eqref{eq:Smoothness} is difficult.
This is because, for a typical neural network architecture, it is unclear in practice if the function $[\bm{x} - \bm{\psi}(\cdot)]^2_a$ has {$L_a$}-Lipschitz continuous Hessian for all $a\in [d]$. Hence we propose below another AE loss (splitted loss) in order to cancel the third term in~\eqref{eq:Smoothness}.
Nevertheless, the assumption in Lemma~\ref{Lemma:Smoothness} is used to provide a meaningful bound on the remainder in~\eqref{eq:Smoothness}. In the light of these remarks, we propose  two stochastic AE losses as described below.
\paragraph{AE losses: \label{sec:AE_losses}}
In analogy with the VAE objective~\eqref{eq:VAE_AE}, the first AE encoder loss function can be chosen as
\begin{equation*}
	L^{(\sigma)}_{\bm{\xi},\mathbb{P}_{U}}(\bm{x},\bm{z}) =\mathbb{E}_{\bm{\epsilon}\sim\mathcal{N}(0,\mathbb{I}_{m} )} \left\|\bm{x} - \bm{\psi}_{\bm{\xi}}\big(\mathbb{P}_U\bm{z}+\sigma  U\bm{\epsilon}\big)\right\|_2^2, \text{ with } \sigma>0. \label{eq:AE_VAE}
\end{equation*}
\begin{figure}[h!]
	\centering
	\setlength{\tabcolsep}{2pt}
	\resizebox{0.90\textwidth}{!}{
		\begin{tabular}{r c c c}
			                                                                                          & {\Huge Dsprites} & \Huge 3Dshapes & \Huge Cars3D \\
			\rotatebox[origin=c]{90}{\Huge ${U}_{\star} \quad (\St\text{-RKM-sl}) $}                  &
			\tabfigure{trim={2cm 0.5cm 1.5cm 1.4cm}, clip}{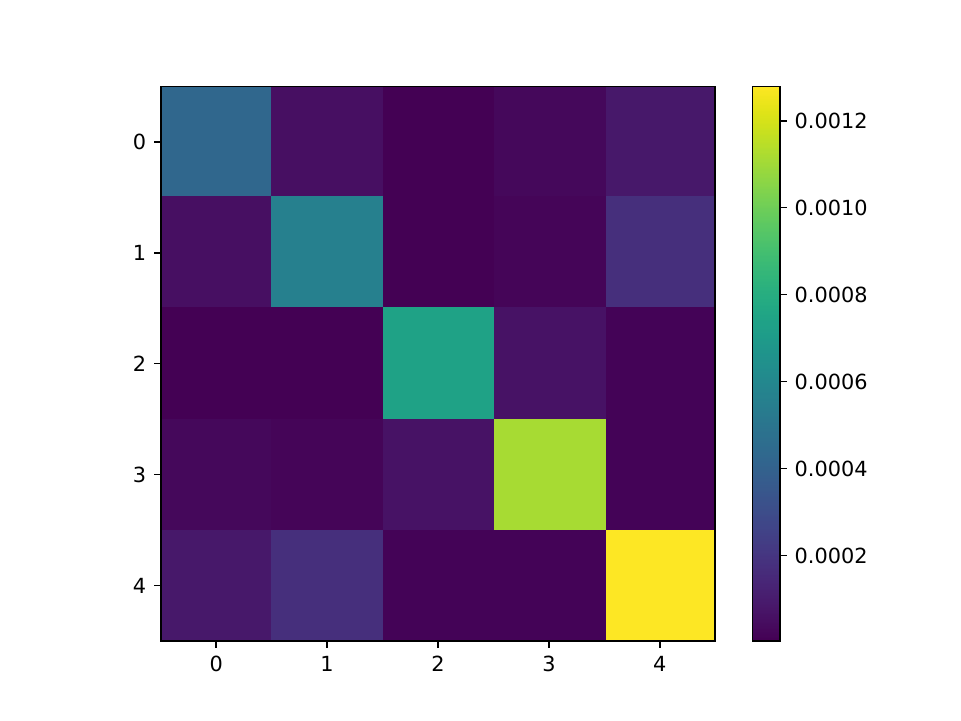}        &
			\tabfigure{trim={2cm 0.5cm 1.5cm 1.4cm}, clip}{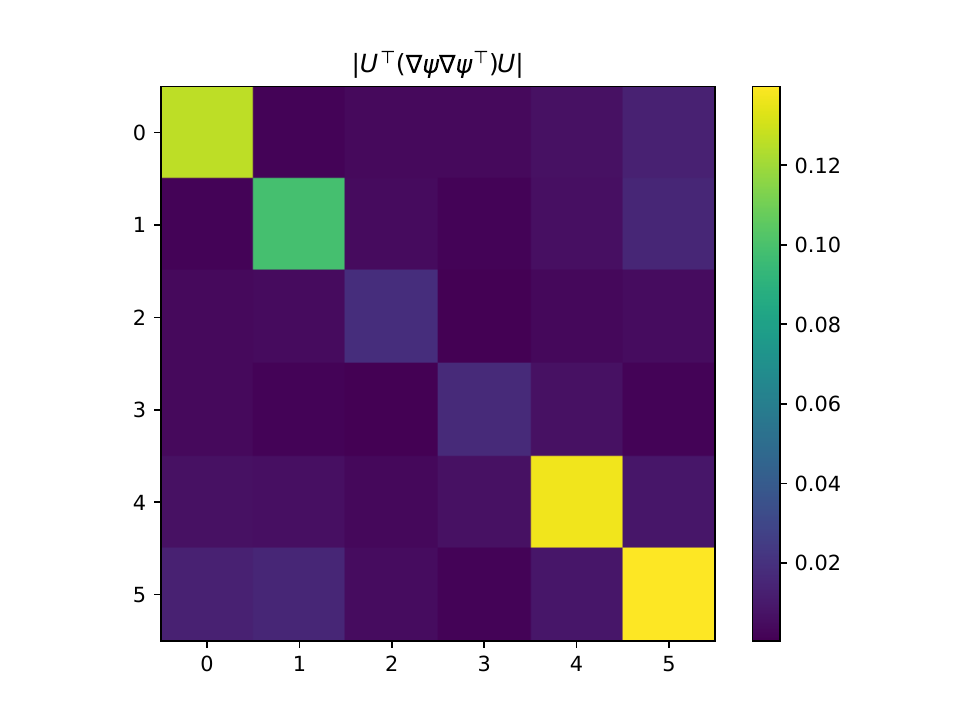}        &
			\tabfigure{trim={1.6cm 0.5cm 1.5cm 1.3cm}, clip}{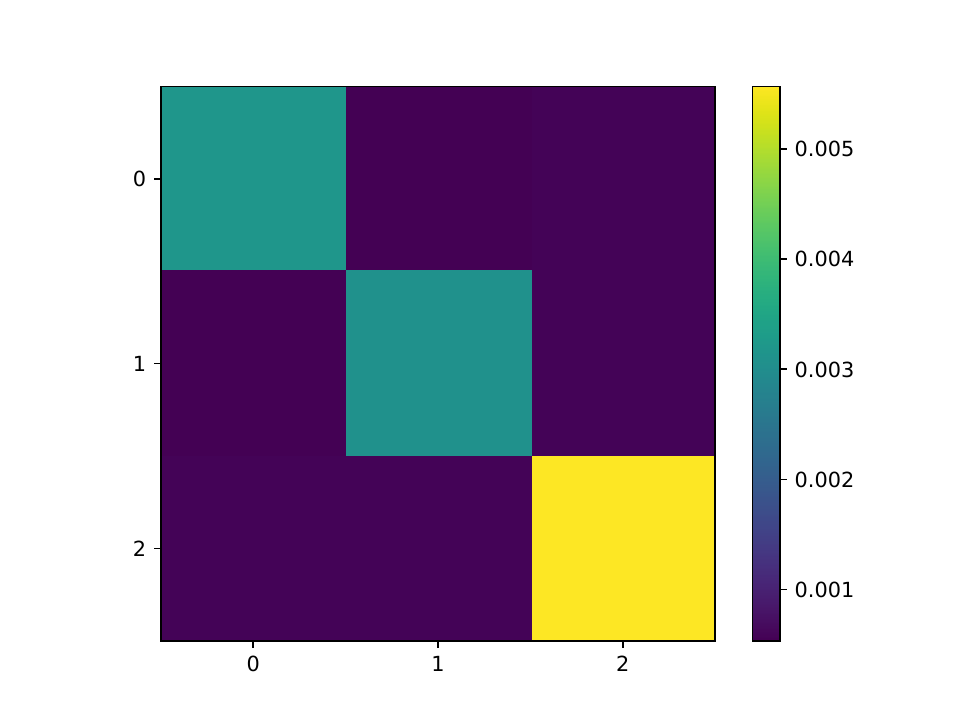}                                                           \\
			\rotatebox[origin=c]{90}{\Huge ${U}_{\star} \quad (\St\text{-RKM}) $}                     &
			\tabfigure{trim={2cm 0.5cm 1.5cm 1.4cm}, clip}{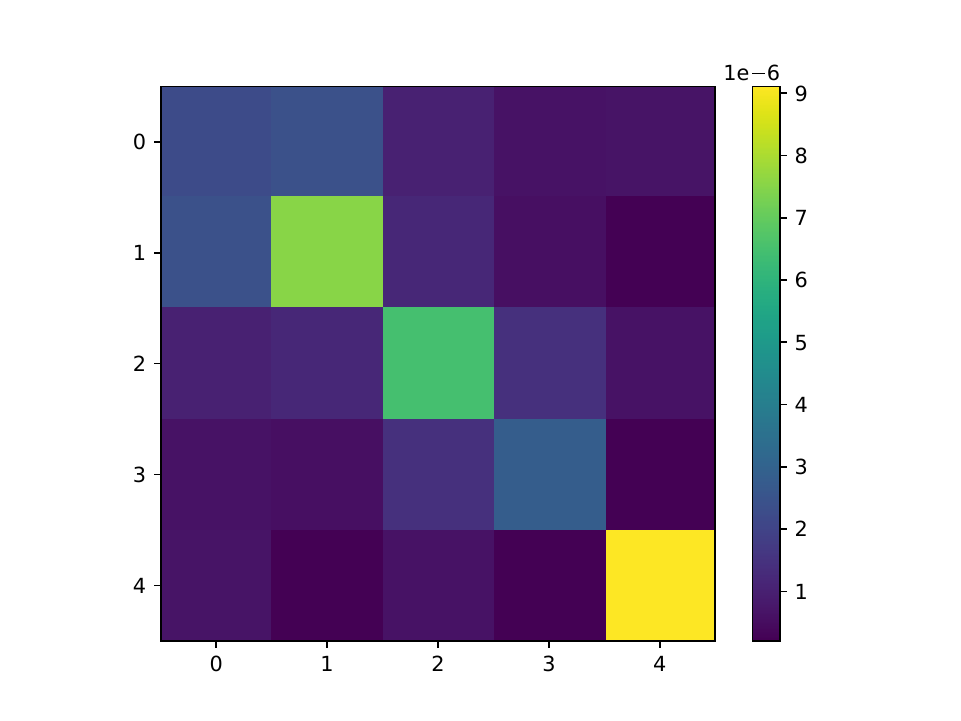} &
			\tabfigure{trim={2cm 0.5cm 1.5cm 1.4cm}, clip}{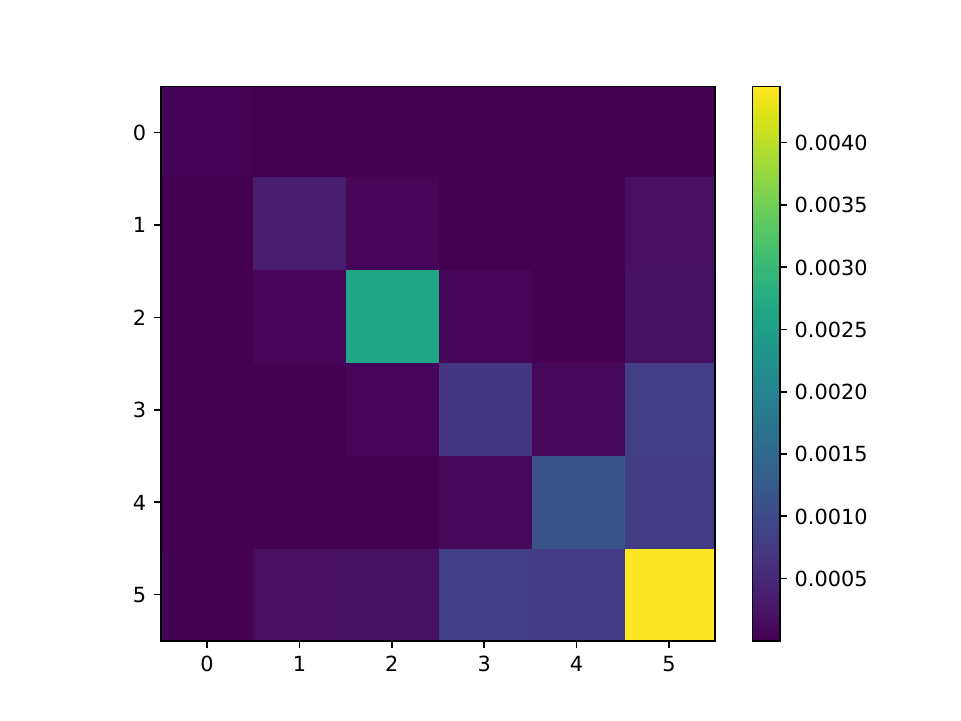} &
			\tabfigure{trim={1.6cm 0.5cm 1.5cm 1.3cm}, clip}{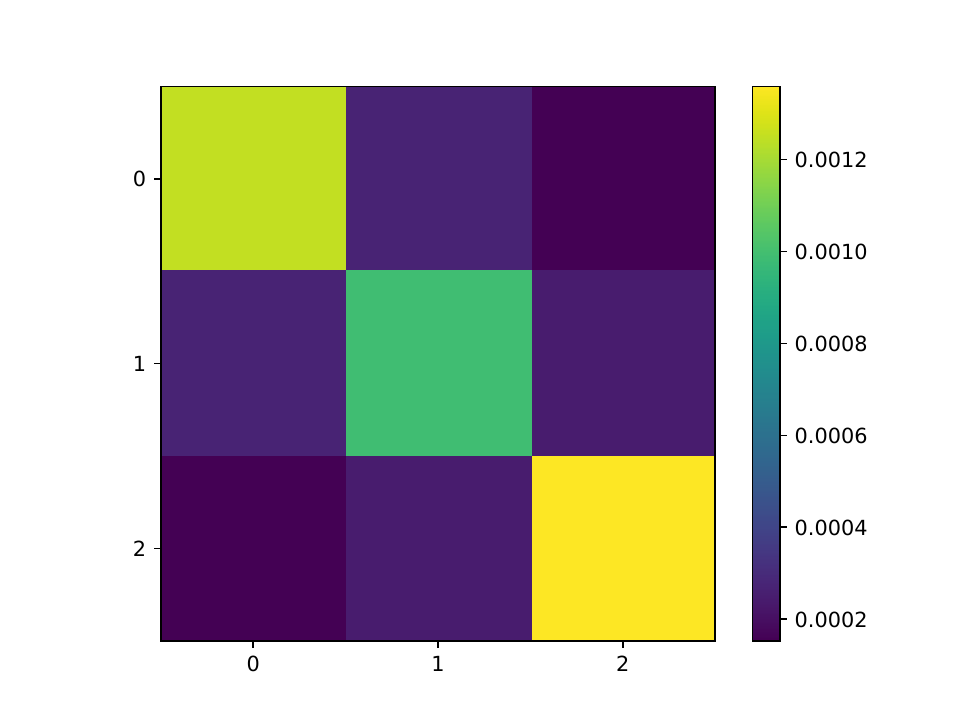}                                                    \\
			\rotatebox[origin=c]{90}{\Huge Random ${U}$}                                              &
			\tabfigure{trim={2cm 0.5cm 1.5cm 1.4cm}, clip}{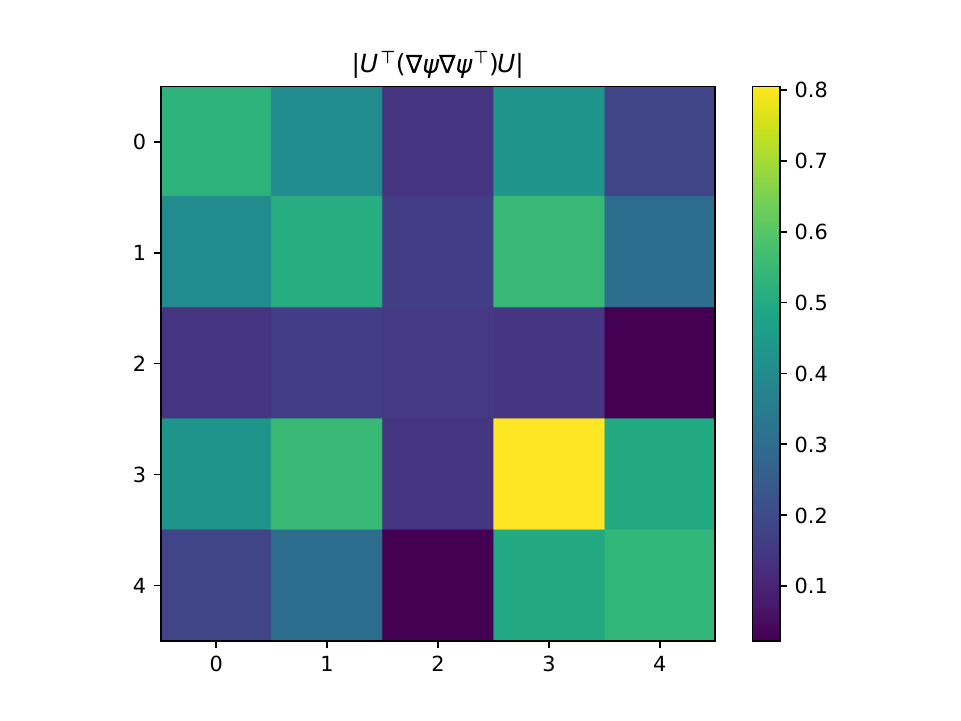} &
			\tabfigure{trim={2cm 0.5cm 1.5cm 1.4cm}, clip}{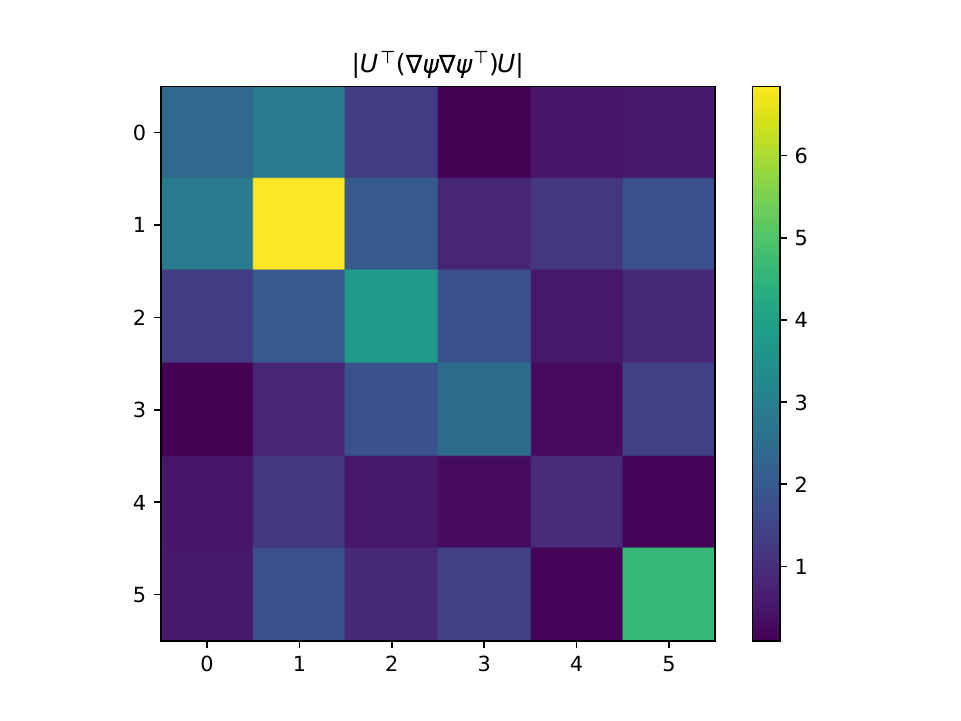} &
			\tabfigure{trim={1.6cm 0.5cm 1.5cm 1.3cm}, clip}{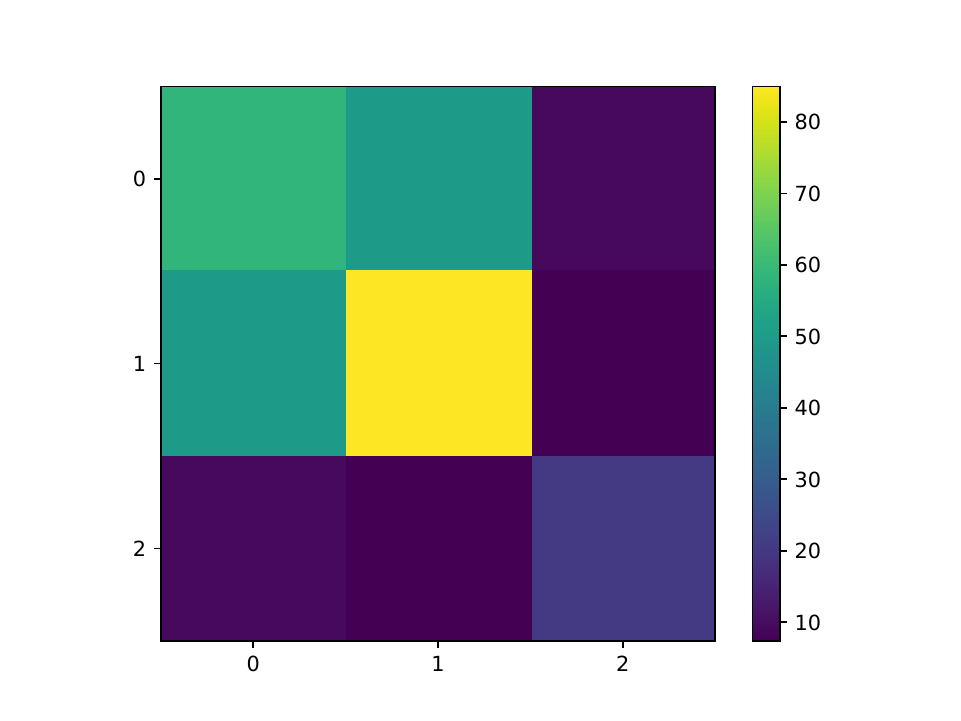}
		\end{tabular}
	}
	\caption{Visualizing the matrix~\eqref{eq:MC_covariance_of_variations} for St-RKM models after training on three datasets. The first two rows show~\eqref{eq:MC_covariance_of_variations} where $U = {U}_{\star} \in \St(\ell,m)$ is the output of Algorithm~\ref{algo}. These matrices are effectively close to being diagonal and especially for $\St\text{-RKM-sl}$, as it is expected.  In contrast, the third row shows the same matrix~\eqref{eq:MC_covariance_of_variations} with ${U}\in \St(\ell,m)$ sampled uniformly at random; see Table~\ref{Table:diag_scores} for the corresponding normalized diagonalization errors.}
	\label{fig:resultant_matrices}
\end{figure}
As motivated by Lemma~\ref{Lemma:Smoothness} above, the noise term $\sigma U\bm{\epsilon}$ above promotes a \emph{smoother} decoder network. To further promote disentanglement, we propose a splitted AE loss
\begin{equation}
	L^{(\sigma),sl}_{\bm{\xi},\mathbb{P}_{U}}(\bm{x},\bm{z}) =\left\|\bm{x} - \bm{\psi}_{\bm{\xi}}\big(\mathbb{P}_U\bm{z}\big)\right\|_2^2 \\ +\mathbb{E}_{\bm{\epsilon}} \left\|\bm{\psi}_{\bm{\xi}}\big(\mathbb{P}_U\bm{z}\big) - \bm{\psi}_{\bm{\xi}}\big(\mathbb{P}_U\bm{z}+\sigma  U\bm{\epsilon}\big)\right\|_2^2,\label{eq:slLoss}
\end{equation}
with $\bm{\epsilon}\sim\mathcal{N}(0,\mathbb{I}_m)$.
The first term in~\eqref{eq:slLoss} is the classical AE loss while the second term promotes orthogonal directions of variations.
Thus, by relating Lemma~\ref{Lemma:Smoothness} to~\eqref{eq:slLoss}, we see that
\begin{equation*}
	L^{(\sigma),sl}_{\bm{\xi},\mathbb{P}_{U}}(\bm{x},\bm{z}) =\left\|\bm{x} - \bm{\psi}_{\bm{\xi}}\big(\mathbb{P}_U\bm{z}\big)\right\|_2^2 \\ +\sigma^2  \Tr\big(U^\top\nabla\bm{\psi}(\bm{y})\nabla\bm{\psi}(\bm{y})^\top U\big) \\
	+ \sum_{a=1}^d R_a(\sigma).
\end{equation*}
In short,
the optimization over $U$ in~\eqref{eq:ReducedObjective} with the splitted loss aims to promote a $U_\star$ such that
\begin{center}
	$  U^\top_\star C_{\bm{\theta}} U_\star $ and $ U^\top_\star\left(\sum_{i=1}^n\nabla\bm{\psi}(\bm{y}_i)\nabla\bm{\psi}(\bm{y}_i)^\top \right)U_\star $ are \emph{jointly diagonal}.
\end{center}
Figure~\ref{fig:resultant_matrices} gives a visualization of the diagonal form of
\begin{equation}
	\frac{1}{|\mathcal{C}|}\sum_{i\in \mathcal{C}}U^\top_\star\nabla\bm{\psi}(\bm{y}_i)\nabla\bm{\psi}(\bm{y}_i)^\top U_\star, \text{ with } \bm{y}_i = \mathbb{P}_{U}\bm{\phi_{\theta}}(\bm{x}_i)\label{eq:MC_covariance_of_variations}
\end{equation}
obtained after training; where $\mathcal{C}$ contains the indices of a subset of $50$ images sampled uniformly at random. For numerical values, Table~\ref{Table:diag_scores} in the appendix shows the normalized diagonalization errors.

Note that we do not simply propose another encoder-decoder architecture, given by: $U^\top\bm{\phi}_{\bm{\theta}}(\cdot)$ and $\bm{\psi}_{\bm{\xi}}(U \cdot)$.
Instead, our objective assumes that the neural network defining the encoder provides a better embedding if we impose that it maps training points on a linear subspace of dimension $m<\ell$ in the $\ell$-dimensional latent space.
In other words, the optimization of the parameters in the last layer of the encoder does not play a redundant role, since the second term in  \eqref{eq:ReducedObjective} clearly also depends on $\mathbb{P}_{U^\perp}\bm{\phi}_{\bm{\theta}}(\cdot)$. The full training involves an alternating minimization procedure which is described in Algorithm~\ref{algo}.
%
%%%%
\subsection{Contributions~\label{sec:contributions}}
Here is a summary of our contributions. We propose two main changes with respect to the related works: (\emph{i}) To promote disentangled representation learning, we propose orthogonal projection in the latent space via a rectangular matrix which is valued on the Stiefel manifold. Then for the training, we use the Cayley ADAM algorithm of~\cite{Li2020Efficient} for stochastic optimization on the Stiefel manifold and call our proposed model $\St$-RKM.
(\emph{ii})  We propose several objective functions to learn the feature map and the pre-image map  networks in the form of an encoder and a decoder respectively.
The best configuration for promoting a disentangled representation is
\begin{equation*}
	\min_{\substack{ U\in \St(\ell,m)\\\bm{\theta}, \bm{\xi}}} \frac{\lambda}{n}\sum_{i=1}^n \text{(splitted) AE loss}(\bm{x}_i,\mathbb{P}_U,\bm{\theta}, \bm{\xi}) + \text{PCA objective}(C_{\bm{\theta}},\mathbb{P}_U),
\end{equation*}
where the covariance matrix reads $
	C_{\bm{\theta}} = \frac{1}{n}\sum_{i=1}^n  \bm{\phi}_{\bm{\theta}}(\bm{x}_i)\bm{\phi}_{\bm{\theta}}^\top(\bm{x}_i)$ and $\mathbb{P}_U = UU^\top$ with $U$ an $\ell\times m$ matrix with orthonormal columns. Here $\lambda >0$ is a trade-off parameter. The final parameters $(U_\star, \bm{\theta}, \bm{\xi})$  give a local minimizer of this objective with $U_\star$ chosen such that $U^\top_\star C_{\bm{\theta}} U_\star$ is diagonal.
(\emph{iii}) We validate through experiments the following statement:  the combination of a splitted AE loss with a PCA objective by using an explicit optimization on the Stiefel manifold promotes disentanglement. In this paper, disentanglement is interpreted as jointly  diagonalizing  the matrix representing variations in the input space with respect to latent motions $\sum_i U^\top_\star\nabla\bm{\psi_{\xi}}(\bm{y}_i)\nabla\bm{\psi_{\xi}}(\bm{y}_i)^\top U_\star$ where $\bm{y}_i = \mathbb{P}_{U_\star} \bm{\phi}_{\bm{\theta}}(\bm{x}_i)$ and the covariance matrix of the dataset in the latent space $U^\top_\star C_{\bm{\theta}} U_\star$.

\begin{figure}
	\centering
	\setlength\tabcolsep{0pt}
	\def\arraystretch{0.0}%  1 is the default, change whatever you need
	\begin{tabu}{cc}
		\tabfigure{width=0.49\textwidth}{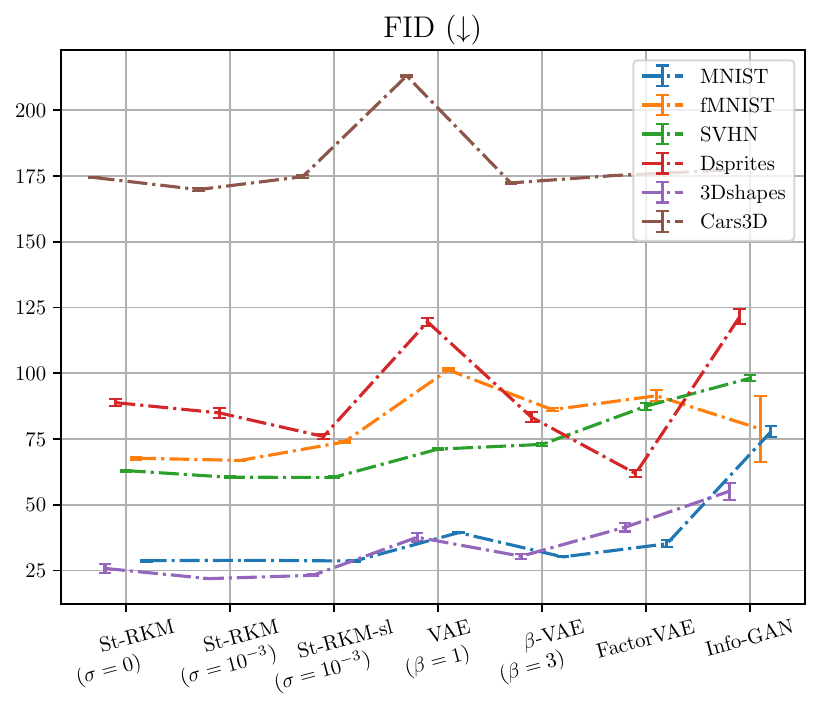} &
		\tabfigure{width=0.49\textwidth}{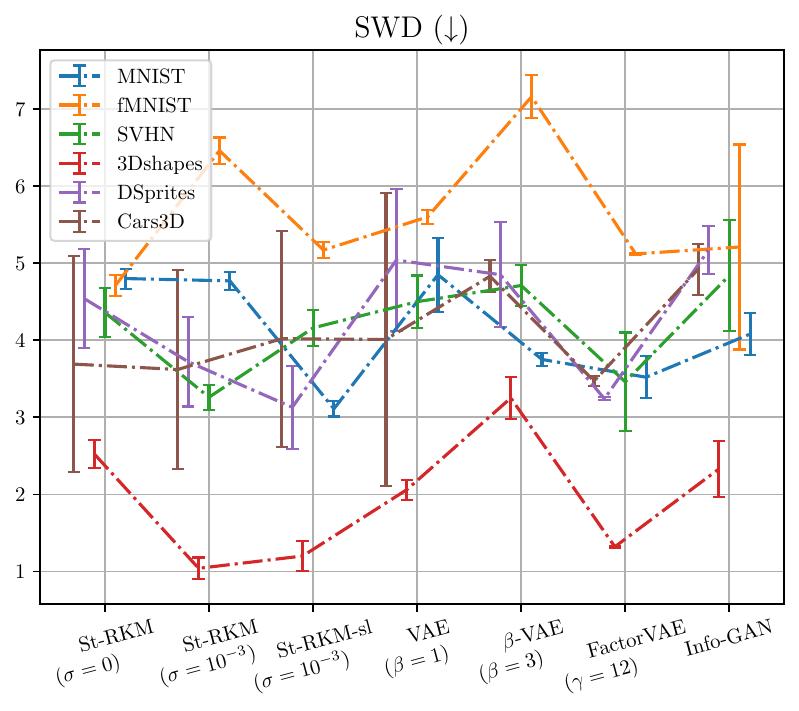}
	\end{tabu}
	\caption{Fr\'echet Inception Distance (FID)~\cite{Heusel2017} and Sliced Wasserstein Distance (SWD) scores (mean and 1 standard-deviation) for 8000 randomly generated samples (smaller is better).}
	\label{fig:fid_swd}
\end{figure}
\section{Connections with the Evidence Lower Bound \label{sec:elbo}}%

We now discuss the interpretation of the proposed model in the probabilistic setting and the independence of latent factors. In order to formulate an ELBO, consider the following random encoders:
\begin{equation*}
	q(\bm{z}|\bm{x})=\mathcal{N}(\bm{z}|\bm{\phi}_{\bm{\theta}}(\bm{x}),\gamma^2\mathbb{I}_\ell) \text{ and }  q_U(\bm{z}|\bm{x})=\mathcal{N}(\bm{z}|\mathbb{P}_U\bm{\phi}_{\bm{\theta}}(\bm{x}),\sigma^2\mathbb{P}_U+\delta^2 \mathbb{P}_{U^\perp}),
\end{equation*}
where $\bm{\phi}_{\bm{\theta}}$ has zero mean on the data distribution. Here, $\sigma^2$ plays the role of a trade-off parameter, while the regularization parameter $\delta$ is introduced for technical reasons and is put to a numerically small absolute value (see Appendix for details). Let the decoder be $p(\bm{x}|\bm{z}) =  \mathcal{N}(\bm{x}|\bm{\psi_{\xi}}(\bm{z}),\sigma^2_0\mathbb{I})$ and the latent space distribution is parametrized by $p(\bm{z}) = \mathcal{N}(0,\Sigma)$ where $\Sigma\in \mathbb{R}^{\ell\times\ell}$ is a covariance matrix.
We treat $\Sigma$ as a parameter of the optimization problem that is determined at the last stage of the training.
Then the minimization problem \eqref{eq:ReducedObjective} with stochastic AE loss is equivalent to the maximization of
\begin{equation}
	\frac{1}{n}\sum_{i=1}^{n}\Big\{\underbrace{\mathbb{E}_{ q_U(\bm{z}|\bm{x}_i)}[\log (p(\bm{x}_i|\bm{z}))]}_{\text{(I)}} -  \underbrace{\KL(q_U(\bm{z}|\bm{x}_i),q(\bm{z}|\bm{x}_i))}_{\text{(II)}} - \underbrace{\KL(q_{U}(\bm{z}|\bm{x}_i),p(\bm{z}))}_{\text{(III)}}\Big\},\label{eq:ProbaRKM0}
\end{equation}
which is a lower bound to the ELBO, since the KL divergence in (II) is positive. {For details of the derivation, see Appendix.} The hyper-parameters $\gamma,\sigma,\sigma_0$ take a fixed value.
Up to additive constants, the terms (I) and (II) of \eqref{eq:ProbaRKM0} match the objective \eqref{eq:ReducedObjective}. The third term (III) in \eqref{eq:ProbaRKM0} is optimized after the training of the first two terms. It can be written as follows
\begin{equation*}
	\frac{1}{n}\sum_{i=1}^{n}\KL( q_{U}(\bm{z}|\bm{x}_i),p(\bm{z}))=\frac{1}{2}\Tr [\Sigma_0\Sigma^{-1}] \\+\frac{1}{2}  \log(\det \Sigma) + \text{constants},\label{eq:KLlatent}
\end{equation*}
with $\Sigma_0 =\mathbb{P}_{U}C_{\bm{\theta}}\mathbb{P}_{U}+\sigma^2\mathbb{P}_U+\delta^2 \mathbb{P}_{U^\perp}$.
Hence, in that case, the optimal covariance matrix is diagonalized $ \Sigma = U(\diag(\bm{\lambda})+\sigma^2\mathbb{I}_{m})U^\top +  \delta^2\mathbb{P}_{U_{\perp}}, $
with $\bm{\lambda}$ denoting the principal values of the PCA.

Now we briefly discuss the factorization of the encoder.
Let $\bm{h}(\bm{x}) = U^\top \bm{\phi}_{\bm{\theta}}(\bm{x})$ and let the `effective' latent variable be $ \bm{z}^{(U)} = U^\top\bm{z}\in \mathbb{R}^m$.
Then the probability density function of $q_U(\bm{z}|\bm{x})$ is
\begin{equation*}
	f_{q_U(\bm{z}|\bm{x})}(\bm{z})= \frac{e^{-\frac{\| U_{\perp}^\top\bm{z}\|_2^2}{2\delta^2}}}{(\sqrt{2\pi\delta^2})^{\ell-m}} \prod_{j=1}^m\frac{e^{-\frac{(\bm{z}^{(U)}_j- \bm{h}_j(\bm{x}))^2}{2\sigma^2}} }{\sqrt{2\pi\sigma^2}},
\end{equation*}
where the first factor is approximated by a Dirac delta if $\delta\to 0$. Hence, the factorized form of $q_U$ shows the independence of the latent variables $\bm{z}^{(U)}$. This factorization is used as a regularization term in the objective by \cite{FactorVAE} to promote disentanglement. In particular, the term (II) in \eqref{eq:ProbaRKM0} is analogous to a `Total Correlation' loss~\cite{MIG_VAE}.

\begin{figure}[h!]
	\centering
	\setlength{\tabcolsep}{2pt}
	\resizebox{\textwidth}{!}{
		\begin{tabular}{r c c}
			                                   & {\normalsize RKM-sl} ($\sigma=10^{-3}$)                            & FactorVAE ($\gamma=12$)                                                  \\
			\rotatebox[origin=c]{90}{3Dshapes} & \tabfigure{width=6.5cm, height=4cm}{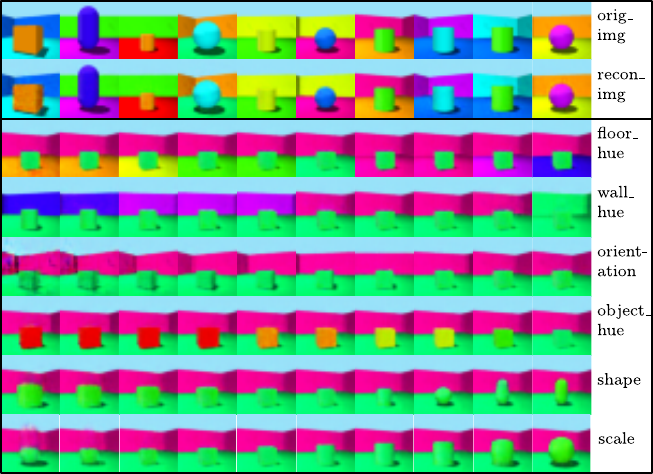}    & \tabfigure{width=6.5cm, height=4cm}{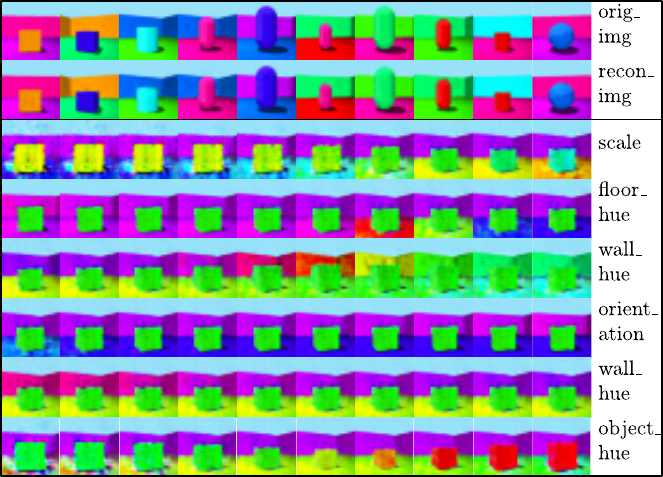}    \\
			\rotatebox[origin=c]{90}{Dsprites} & \tabfigure{width=6.5cm, height=4cm}{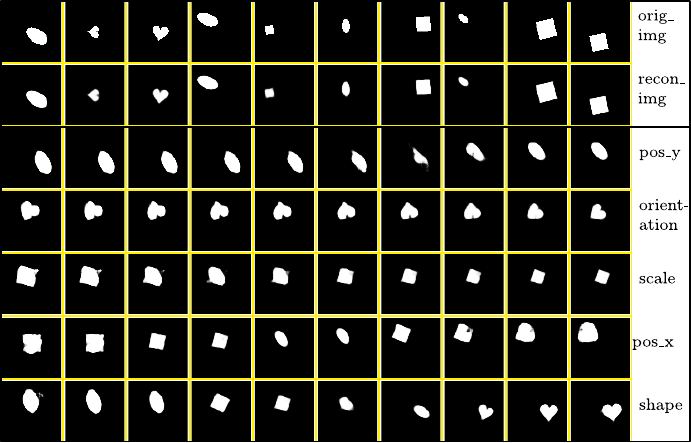}   & \tabfigure{width=6.5cm, height=4cm}{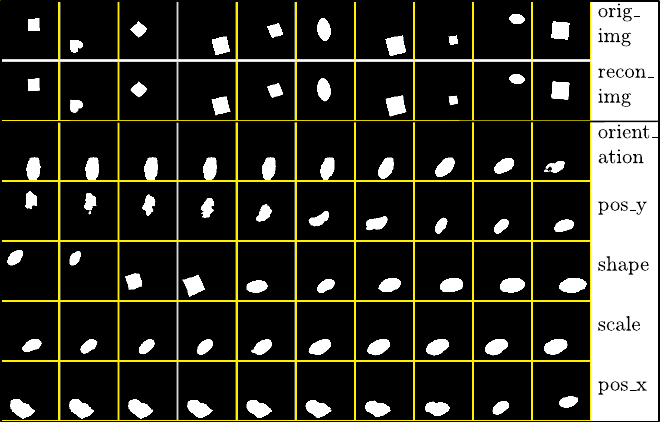}    \\
			\rotatebox[origin=c]{90}{Cars3D}   & \tabfigure{width=6.5cm, height=3.5cm}{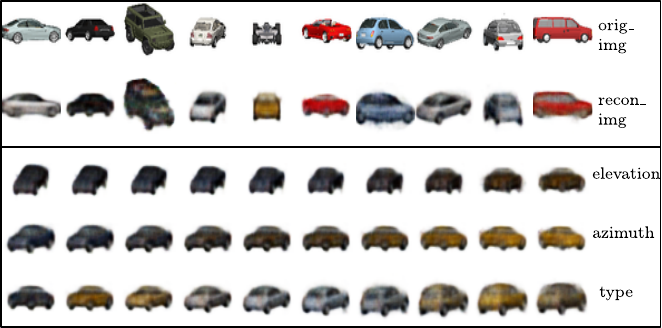} & \tabfigure{width=6.5cm, height=3.5cm}{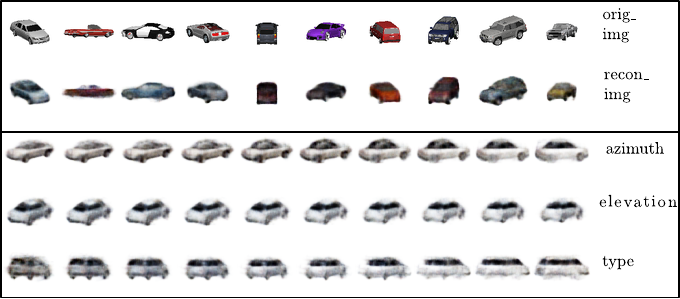}
		\end{tabular}
	}
	\caption{Traversals along the principal components. First-two rows show the ground-truth and reconstructed images. Further, each subsequent row shows the generated images by traversing along a principal component in the latent space.  The last column in each sub-image indicates the dominant factor of variation.}
	\label{fig:traversal_imgs}
\end{figure}
\section{Experiments \label{sec:exp}}
In this section, we investigate if $\St$-RKM\footnote{The source code is available at \href{http://bit.ly/StRKM_code}{http://bit.ly/StRKM\_code}} can simultaneously achieve (i) accurate reconstructions on training data (ii) good random generations, and (iii) good disentanglement performance. We use the standard datasets: MNIST~\cite{lecun-mnisthandwrittendigit2010}, Fashion-MNIST~\cite{xiao2017/online} (fMNIST), and SVHN~\cite{netzerReadingDigitsNatural}. To evaluate disentanglement, we use datasets with known ground-truth generating factors such as Dsprites~\cite{dsprites17}, 3Dshapes~\cite{3dshapes18}, and Cars3D~\cite{reed2015deep}. Further, all figures and tables report average errors with 1 standard deviation over 10 experiments. %
\begin{table}[ht]
	\centering
	\caption{Training time in minutes (for 1000 epochs, mean with 1 standard deviation over 10 runs) and the number of parameters (Nb) of the generative models on the MNIST dataset.}
	\label{tab:train_times}
	\begin{tabular}{lcccc}
		\textbf{Model}         & $\St$-\textbf{RKM} & ($\beta$)-\textbf{VAE} & \textbf{FactorVAE} & \textbf{Info-GAN} \\ \midrule
		\textbf{Nb parameters} & $\bm{4164519}$     & $4165589$              & $8182591$          & $4713478$         \\
		\textbf{Training time} & 21.93 (1.3)        & $\bm{19.83}$~(0.8)     & 33.31 (2.7)        & 45.96 (1.6)
	\end{tabular}
\end{table}

\noindent \textbf{Algorithm}: We use an alternating-minimization scheme as shown in Algorithm~\ref{algo}. First, the Adam optimizer with a learning rate $2\times 10^{-4}$ is used to update the encoder-decoder parameters and then, the Cayley Adam optimizer~\cite{Li2020Efficient} with a learning rate $10^{-4}$ is used to update  $U$. Finally, at the end of the training, we recompute $U$ from the Singular Value Decomposition (SVD) of the covariance matrix as a final correction-step of the Kernel PCA term in our objective (step 10 of Algorithm \ref{algo}). Since the $\ell \times \ell $ covariance matrix is typically small, this decomposition is fast (see Table \ref{tab:arch}). In practice, our training procedure only marginally increases the computation cost which can be seen from training times in Table~\ref{tab:train_times}.

\begin{algorithm}[h]
	\caption{Manifold optimization of $\St$-RKM}\label{algo}
	\textbf{Input:} $ \{\bm{x}_{i}\}_{i=1}^n$, $ \bm{\phi}_{\bm{\theta}}, \bm{\psi}_{\bm{\zeta}}, \mathcal{J}\coloneqq \text{Eq.~} \ref{eq:ReducedObjective} $\\
	\textbf{Output:} Learned $\bm{\theta, \zeta} , U$
	\begin{algorithmic}[1]
		\Procedure{Train}{}
		\While{not converged}
		\State $\{\bm{x}\} \gets \text{\{Get mini-batch\}} $
		\State Get embeddings $ \bm{\phi}_{\bm{\theta}}(\bm{x})\gets \bm{x} $
		\State Compute centered $C_{\bm{\theta}}$\Comment{Covariance matrix}
		\State $ \text{Update}~\{ \bm{\theta}_{\bm{e}},\bm{\psi}_{\bm{g}}\}  \gets \text{Adam}(\mathcal{J})$ \Comment{Optimization step}
		\State $ \text{Update}~\{ U\}  \gets \text{Cayley\_Adam}(\mathcal{J})$\Comment{Optimization step}
		\EndWhile
		\State Do steps 4-5 over whole dataset
		\State $ U \gets \text{SVD}(C_{\bm{\theta}})$\Comment{Equation \eqref{eq:svd}}
		\EndProcedure
	\end{algorithmic}
\end{algorithm}

\noindent \textbf{Experimental setup}: We consider four baselines for comparison: (i) VAE, (ii) $\beta$-VAE, (iii) FactorVAE and (iv) Info-GAN. An ablation study with the Gen-RKM is  shown in the appendix~\ref{sec:ablation_study}.
Extensive experimentation was not computationally feasible since the evaluation and decomposition of kernel matrices scales $ \mathcal{O}(n^{2} ) $ and $ \mathcal{O}(n^{3} ) $ with the dataset size; see the discussion in Section \ref{sec:related_works}.

\noindent \textbf{Inductive biases}:
To be consistent in evaluation, we keep the same encoder (discriminator) and decoder (generator) architecture; and the same latent dimension across the models.
We use convolutional neural networks due to the choice of image datasets for evaluating generation and disentanglement.
In the case of Info-GAN, batch-normalization is added for training stability; see appendix~\ref{sec:hyperparams} for details.
For the determination of the hyperparameters of other models, we start from values in the range of the parameters suggested in the authors' reference implementation.
After trying various values we noticed that $\beta = 3$ and $\gamma = 12$ seem to work well across the datasets that we considered for $\beta$-VAE and FactorVAE respectively.
Furthermore, in all the experiments on $\St$-RKM, we keep the reconstruction weight $\lambda=1$.
All models are trained on the entire dataset.
Note that for the same encoder-decoder network, the $\St$-RKM model has the least number of parameters compared to any VAE variants and Info-GAN (see Table~\ref{tab:train_times}).

To evaluate the quality of generated samples, we report the Fr\'echet Inception Distance~\cite{Heusel2017} (FID)  and the Sliced Wasserstein Distance (SWD)~\cite{karras2017progressive} scores with mean and standard deviation in Figure \ref{fig:fid_swd}. Note that FID scores are not necessarily appropriate for Dsprites since this dataset is significantly different from ImageNet on which the Inception network was originally trained. Randomly generated samples are shown in Figure~\ref{fig:fid_sample_imgs} in Appendix.
To generate samples from the deterministic $\St$-RKM ($\sigma=0$), we sample from a fitted normal distribution on the latent embedding of the dataset; for a similar prodedure, see~\cite{ghosh2019variational}).
Figure~\ref{fig:fid_swd} shows that the $\St$-RKM variants  perform better (lower mean scores) on most datasets and within them, the stochastic variants with $\sigma= 10^{-3}$ perform best.
This can be attributed to a better generalization of the decoder network due to the addition of noise-term on latent-variables; see Lemma~\ref{Lemma:Smoothness}.
The training times for $\St$-RKM variants are shorter compared to FactorVAE and Info-GAN due to a significantly small number of parameters.

\begin{figure}
	\centering
	\setlength\tabcolsep{0pt}
	\def\arraystretch{0.0}%  1 is the default, change whatever you need
	\begin{tabu}{cccc}
		\rotatebox{90}{\scriptsize Lasso}                                           &
		\tabfigure{width=0.32\textwidth}{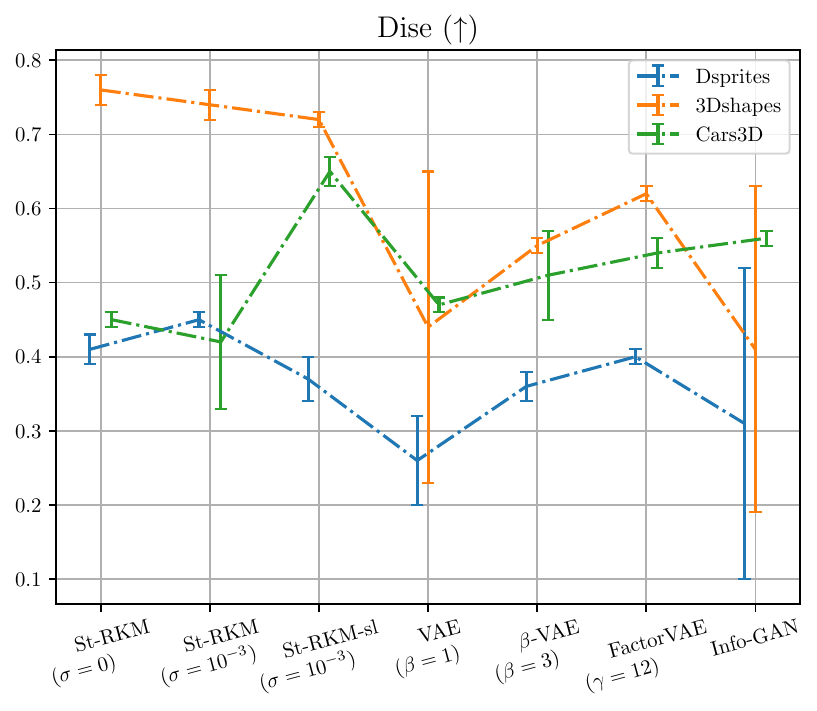} &
		\tabfigure{width=0.32\textwidth}{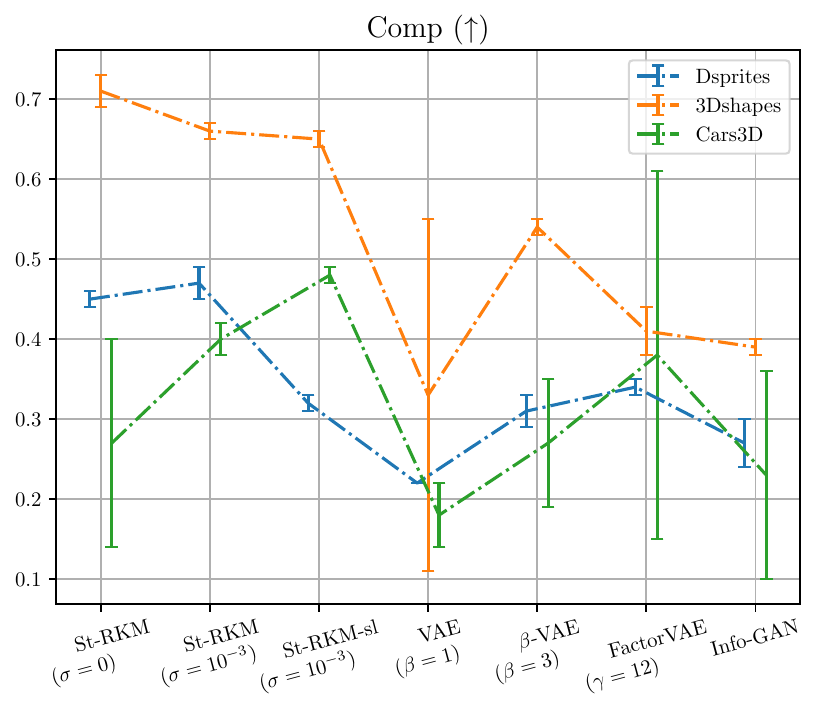} &
		\tabfigure{width=0.32\textwidth}{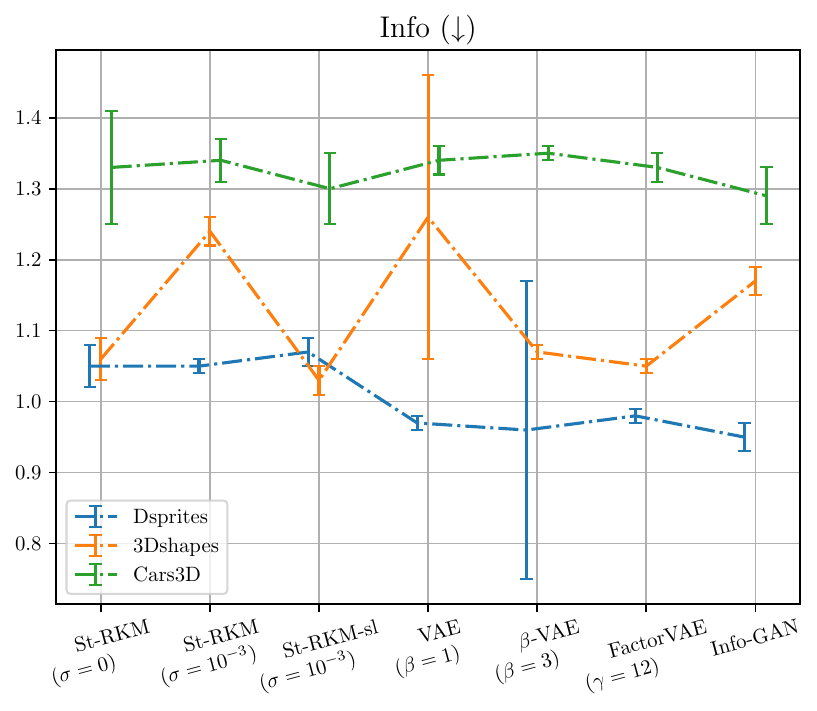}   \\
		\rotatebox{90}{\scriptsize RF}                                              &
		\tabfigure{width=0.32\textwidth}{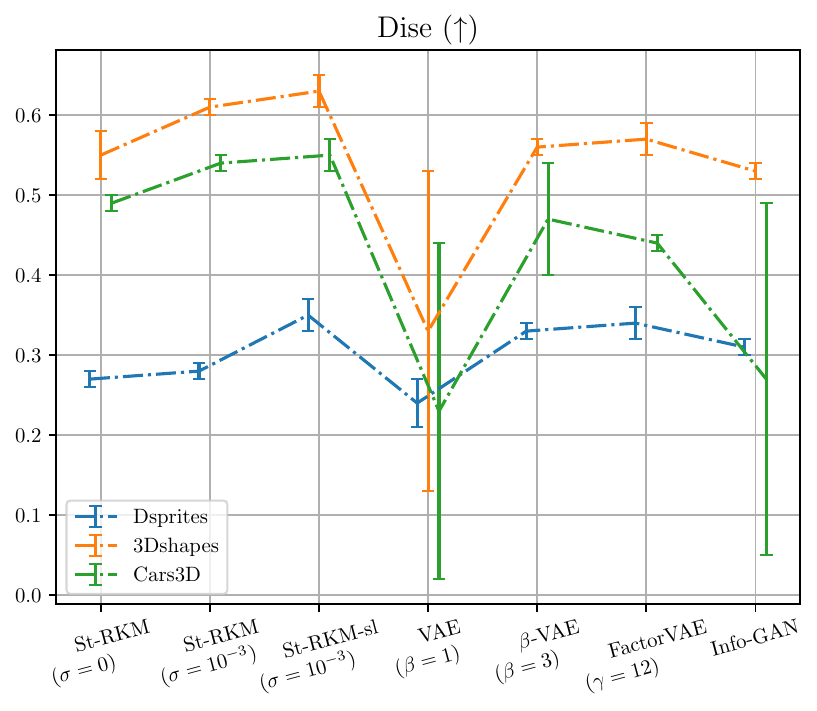}    &
		\tabfigure{width=0.32\textwidth}{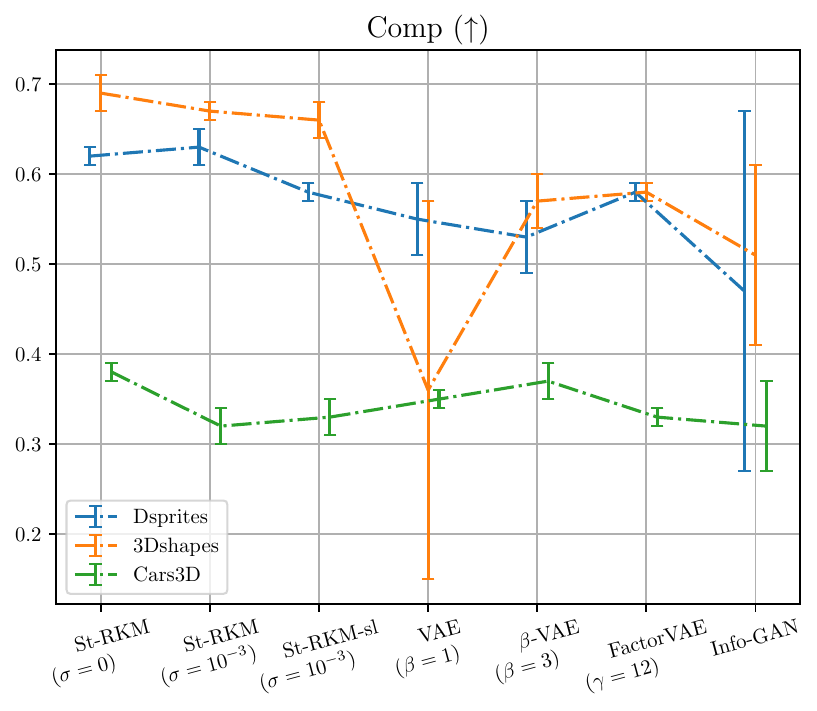}    &
		\tabfigure{width=0.32\textwidth}{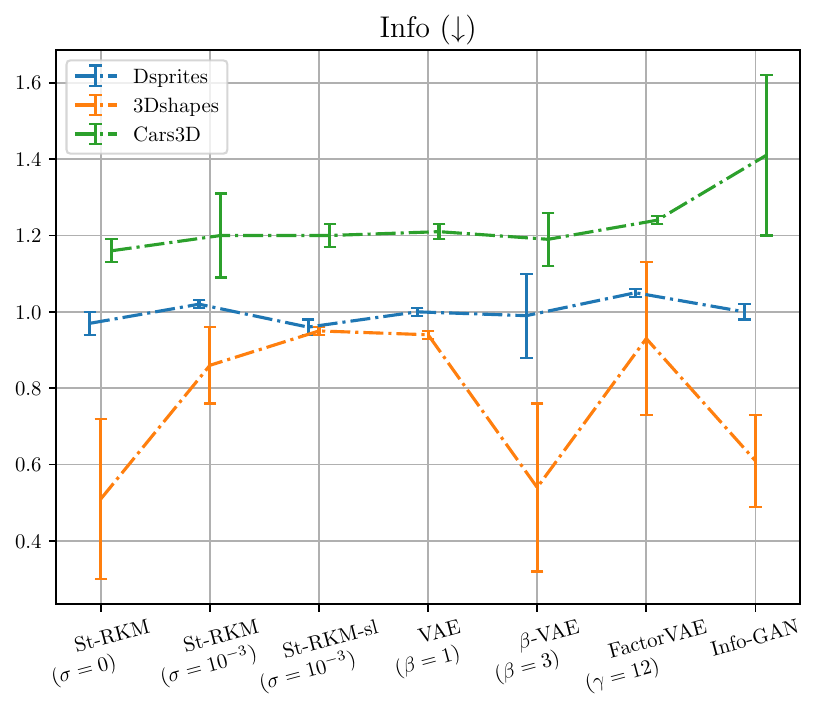}
	\end{tabu}
	\caption{Eastwood framework's~\cite{eastwood2018a} disentanglement metric with Lasso and Random Forest (RF) regressor. The plot shows mean and 1 standard-deviation of scores over 10 iterations. For disentanglement and completeness higher score is better, for informativeness, lower is better. `Info. indicates (average) root-mean-square error in predicting $\bm z$.\label{tab:eastwood}}
\end{figure}

\begin{figure}
	\centering
	\setlength\tabcolsep{0pt}
	\def\arraystretch{0.0}%  1 is the default, change whatever you need
	\begin{tabu}{cc}
		\tabfigure{width=0.49\textwidth}{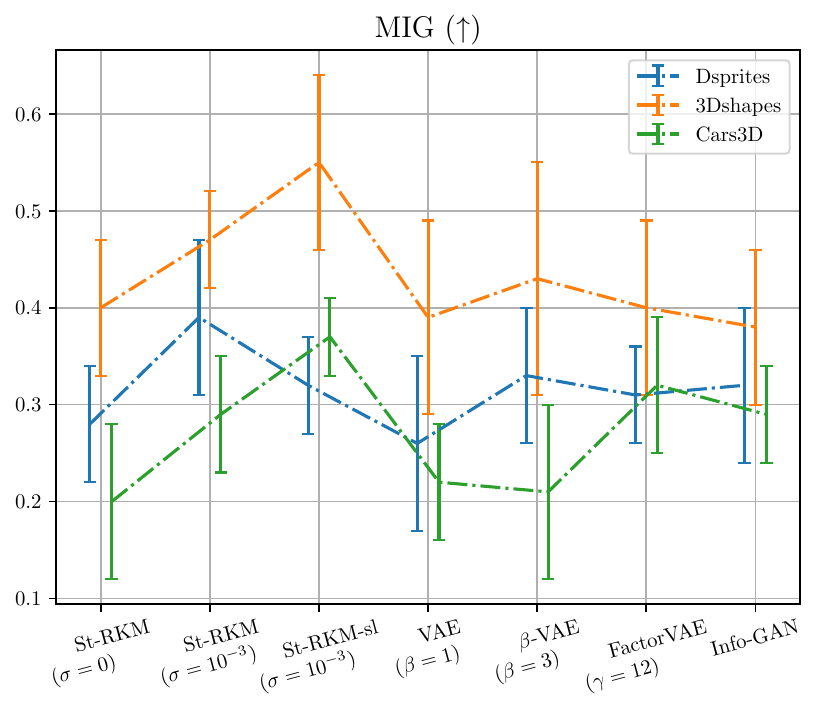} &
		\tabfigure{width=0.49\textwidth}{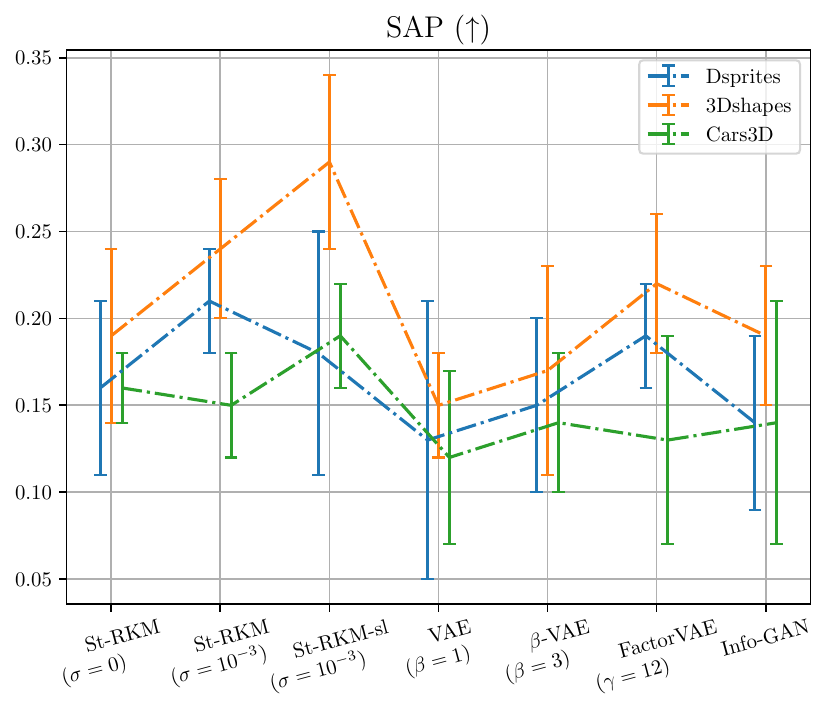}
	\end{tabu}
	\caption{MIG~\cite{MIG_VAE,locatelloChallengingassumptions} and SAP~\cite{kumar2018variational} scores to evaluate disentanglement performance showing the mean (standard deviation) over 10 random seeds.}
	\label{fig:mig_sap}
\end{figure}

To evaluate the disentanglement performance, various metrics have been proposed. A comprehensive review by~\cite{locatelloChallengingassumptions} shows that the various disentanglement metrics are correlated albeit with a different degree of correlation across datasets.
In this paper, we use three metrics to evaluate disentanglement, namely, Eastwood's framework~\cite{eastwood2018a}, Mutual Information Gap (MIG)~\cite{MIG_VAE} and Separated Attribute Predictability (SAP)~\cite{kumar2018variational} scores.
Eastwood's framework~\cite{eastwood2018a} further proposes three metrics: \emph{disentanglement}: the degree to which a representation factorizes the underlying factors of variation, with each variable capturing at most one generative factor; \emph{completeness}: the degree to which each underlying factor is captured by a single code variable; and \emph{informativeness}: the amount of information that a representation captures about the underlying factors of variation. Further we use a slightly modified version of MIG score as proposed by \cite{locatelloChallengingassumptions}.
Figure~\ref{tab:eastwood} shows that $\St$-RKM variants  have better disentanglement and completeness scores (higher mean scores).
However, the informativeness scores are higher for $\St$-RKM when using a lasso-regressor in contrast to mixed scores with a Random forest regressor.
Figure~\ref{fig:mig_sap} further complements these observations by showing MIG and SAP scores. Here the $ \St $-RKM-sl model has the highest mean scores for every dataset.
Qualitative assessment can be done from Figure~\ref{fig:traversal_imgs} which shows the generated images by traversing along the principal components in the latent space. In the 3Dshapes dataset, the $\St$-RKM model captures floor-hue, wall-hue, and orientation perfectly but has a slight entanglement in capturing other factors. This is worse in $\beta$-VAE which has entanglement in all dimensions except the floor-hue along with noise in some generated images.
Similar trends can be observed in the Dsprites and Cars3D datasets.

\section*{Conclusion} This paper proposes $\St$-RKM model for disentangled representation learning and generation based on manifold optimization.
For the training, we use the Cayley Adam algorithm of~\cite{Li2020Efficient} for stochastic optimization on the Stiefel manifold. Computationally, $\St$-RKM only increases the training time by a reasonably small amount compared to $\beta$-VAE for instance.
Further, we propose several autoencoder objectives
and discuss that the combination of a stochastic AE loss with an explicit optimization on the Stiefel manifold promotes disentanglement.
Additionally, we establish connections with probabilistic models, formulate an Evidence Lower Bound, and discuss  the independence of latent factors. Where the considered baselines have a trade-off between  generation quality and disentanglement, we improve on both these aspects as illustrated through various experiments. Some limitations of the proposed model are as follows. A first limitation is hyperparameter selection: the number of components in the KPCA, neural network architecture and the final size of the feature map. When additional knowledge on the data is available, we suggest that the user selects the number of components close to the number of underlying generating factors. The final size of the feature map should be large enough so that KPCA extracts meaningful components. Second, we interpret the disentanglement as the two orthogonal changes in the latent space correspond to two orthogonal changes in input space. Although not perfect, we believe this is a reasonable mathematical approximation of the loosely defined notion of `disentanglement'. Moreover, experimental results  confirm this assumption. Among the possible regularizers on the hidden features, the model associated to the squared Euclidean norm was analyzed in detail, while a deeper study of other regularizers is a prospect for further research, in particular for the case of spherical units.

\subsection*{Acknowledgments}
Most of this work was done when Micha\"el Fanuel was at KU Leuven.
EU: The research leading to these results has received funding from
the European Research Council under the European Union's Horizon
2020 research and innovation program / ERC Advanced Grant E-DUALITY
(787960). This paper reflects only the authors' views and the Union
is not liable for any use that may be made of the contained information.
Research Council KUL: Optimization frameworks for deep kernel machines C14/18/068
Flemish Government:
FWO: projects: GOA4917N (Deep Restricted Kernel Machines:
Methods and Foundations), PhD/Postdoc grant
Impulsfonds AI: VR 2019 2203 DOC.0318/1QUATER Kenniscentrum Data
en Maatschappij
This research received funding from the Flemish Government (AI Research Program). The authors are affiliated to Leuven.AI - KU Leuven institute for AI, B-3000, Leuven, Belgium.
Ford KU Leuven Research Alliance Project KUL0076 (Stability analysis
and performance improvement of deep reinforcement learning algorithms)s), ICT 48 TAILOR, Leuven.AI Institute. The computational resources and services used in this work were provided by the VSC (Flemish Supercomputer Center), funded by the Research Foundation - Flanders (FWO) and the Flemish Government department EWI.

\FloatBarrier

\bibliographystyle{plain}
\bibliography{bibliography}

\begin{thebibliography}{10}

\bibitem{AbsilBook}
P.-A. Absil, R.~Mahony, and R.~Sepulchre.
\newblock {\em {Optimization Algorithms on Matrix Manifolds}}.
\newblock Princeton University Press, Princeton, NJ, 2008.

\bibitem{Woodruff}
Haim Avron, Huy Nguyen, and David Woodruff.
\newblock {Subspace Embeddings for the Polynomial Kernel}.
\newblock In {\em Advances in Neural Information Processing Systems},
  volume~27, pages 2258--2266, 2014.

\bibitem{bengio2013representation}
Yoshua Bengio, Aaron Courville, and Pascal Vincent.
\newblock {Representation Learning: A Review and New Perspectives}.
\newblock {\em {IEEE Transactions on Pattern Analysis and Machine
  Intelligence}}, 35(8):1798--1828, 2013.

\bibitem{3dshapes18}
Chris Burgess and Hyunjik Kim.
\newblock {3Dshapes Dataset}.
\newblock https://github.com/deepmind/3dshapes-dataset/, 2018.

\bibitem{burgess2018understanding}
Christopher~P Burgess, Irina Higgins, Arka Pal, Loic Matthey, Nick Watters,
  Guillaume Desjardins, and Alexander Lerchner.
\newblock {U}nderstanding disentangling in $\beta$-{VAE}.
\newblock In {\em {NIPS 2017 Workshop on Learning Disentangled Representations:
  from Perception to Control}}, 2017.

\bibitem{MIG_VAE}
Ricky T.~Q. Chen, Xuechen Li, Roger~B Grosse, and David~K Duvenaud.
\newblock {Isolating Sources of Disentanglement in Variational Autoencoders}.
\newblock In {\em Advances in Neural Information Processing Systems 31}, pages
  2610--2620, 2018.

\bibitem{dupont2018learning}
Emilien Dupont.
\newblock Learning disentangled joint continuous and discrete representations.
\newblock In {\em Proceedings of the 32nd International Conference on Neural
  Information Processing Systems}, pages 708--718, 2018.

\bibitem{eastwood2018a}
Cian Eastwood and Christopher K.~I. Williams.
\newblock {A Framework for the Quantitative Evaluation of Disentangled
  Representations}.
\newblock In {\em {proceedings of the International Conference on Learning
  Representations (ICLR)}}, 2018.

\bibitem{ghosh2019variational}
Partha Ghosh, Mehdi~SM Sajjadi, Antonio Vergari, Michael Black, and Bernhard
  Sch{\"o}lkopf.
\newblock {From Variational to Deterministic Autoencoders}.
\newblock In {\em {proceedings of the International Conference on Learning
  Representations (ICLR)}}, 2020.

\bibitem{Heusel2017}
Martin Heusel, Hubert Ramsauer, Thomas Unterthiner, Bernhard Nessler, and Sepp
  Hochreiter.
\newblock {{GAN}s Trained by a Two Time-scale Update Rule Converge to a Local
  {Nash} Equilibrium}.
\newblock In {\em Advances in Neural Information Processing Systems}, pages
  6629--6640, 2017.

\bibitem{higgins2017beta}
Irina Higgins, Loic Matthey, Arka Pal, Christopher Burgess, Xavier Glorot,
  Matthew Botvinick, Shakir Mohamed, and Alexander Lerchner.
\newblock {{Beta-VAE}: Learning Basic Visual Concepts with a Constrained
  Variational Framework}.
\newblock In {\em {proceedings of the International Conference on Learning
  Representations (ICLR)}}, volume~2, page~6, 2017.

\bibitem{hinton2005}
Geoffrey~E. Hinton.
\newblock What kind of a graphical model is the brain?
\newblock In {\em Proceedings of the 19th International Joint Conference on
  Artificial Intelligence}, IJCAI'05, page 1765–1775, San Francisco, CA, USA,
  2005. Morgan Kaufmann Publishers Inc.

\bibitem{karras2017progressive}
Tero Karras, Timo Aila, Samuli Laine, and Jaakko Lehtinen.
\newblock {Progressive Growing of {GANs} for Improved Quality, Stability, and
  Variation}.
\newblock In {\em proceedings of the International Conference on Learning
  Representations (ICLR)}, 2017.

\bibitem{FactorVAE}
Hyunjik Kim and Andriy Mnih.
\newblock {Disentangling by Factorising}.
\newblock In {\em proceedings of the Thirty-fifth International Conference on
  Machine Learning (ICML)}, volume~80, pages 2649--2658, 2018.

\bibitem{kingma_auto_encoding_2013}
Diederik~P. Kingma and Max Welling.
\newblock {Auto-Encoding Variational Bayes}.
\newblock In {\em {proceedings of the International Conference on Learning
  Representations (ICLR)}}, 2014.

\bibitem{kumar2018variational}
Abhishek Kumar, Prasanna Sattigeri, and Avinash Balakrishnan.
\newblock {Variational} {Inference} {of} {Disentangled} {Latent} {Concepts}
  {From} {Unlabeled} {Observations}.
\newblock In {\em International Conference on Learning Representations}, 2018.

\bibitem{lecun-mnisthandwrittendigit2010}
Yann LeCun and Corinna Cortes.
\newblock {MNIST} handwritten digit database.
\newblock http://yann.lecun.com/exdb/mnist/, 2010.

\bibitem{lecun_learning_2004}
Yann LeCun, Fu~Jie Huang, and Leon Bottou.
\newblock {Learning Methods for Generic Object Recognition with Invariance to
  Pose and Lighting}.
\newblock In {\em {Computer Vision and Pattern Recognition (CVPR)}}, 2004.

\bibitem{Li2020Efficient}
Jun Li, Fuxin Li, and Sinisa Todorovic.
\newblock {Efficient Riemannian Optimization on the Stiefel Manifold via the
  Cayley Transform}.
\newblock In {\em {proceedings of the International Conference on Learning
  Representations (ICLR)}}, 2020.

\bibitem{locatelloChallengingassumptions}
Francesco Locatello, Stefan Bauer, Mario Lučić, Gunnar Rätsch, Sylvain
  Gelly, Bernhard Schölkopf, and Olivier~Frederic Bachem.
\newblock {Challenging Common Assumptions in the Unsupervised Learning of
  Disentangled Representations}.
\newblock In {\em International Conference on Machine Learning (ICML)}, 2019.

\bibitem{Locatello2020Disentangling}
Francesco Locatello, Michael Tschannen, Stefan Bauer, Gunnar Rätsch, Bernhard
  Schölkopf, and Olivier Bachem.
\newblock {Disentangling Factors of Variations Using Few Labels}.
\newblock In {\em { International Conference on Learning Representations
  (ICLR)}}, 2020.

\bibitem{dsprites17}
Loic Matthey, Irina Higgins, Demis Hassabis, and Alexander Lerchner.
\newblock dsprites: Disentanglement testing sprites dataset.
\newblock https://github.com/deepmind/dsprites-dataset/, 2017.

\bibitem{Nesterov}
Yurii Nesterov.
\newblock {\em Introductory Lectures on Convex Optimization: A Basic Course}.
\newblock Springer Publishing Company, Incorporated, 1st edition, 2014.

\bibitem{netzerReadingDigitsNatural}
Yuval Netzer, Tao Wang, Adam Coates, Alessandro Bissacco, Bo~Wu, and Andrew~Y
  Ng.
\newblock {Reading Digits in Natural Images with Unsupervised Feature
  Learning}.
\newblock In {\em NIPS Workshop on Deep Learning and Unsupervised Feature
  Learning}, 2011.

\bibitem{robust2020}
Arun Pandey, Joachim Schreurs, and Johan A.~K. Suykens.
\newblock {Robust Generative Restricted Kernel Machines} using weighted
  conjugate feature duality.
\newblock In {\em {proceedings of the Sixth International Conference on Machine
  Learning, Optimization, and Data Science (LOD)}}, 2020.

\bibitem{GENRKM}
Arun Pandey, Joachim Schreurs, and Johan~A.K. Suykens.
\newblock Generative restricted kernel machines: A framework for multi-view
  generation and disentangled feature learning.
\newblock {\em Neural Networks}, 135:177 -- 191, 2021.

\bibitem{reed2015deep}
Scott Reed, Yi~Zhang, Yuting Zhang, and Honglak Lee.
\newblock {Deep Visual Analogy-Making}.
\newblock In {\em Advances in Neural Information Processing Systems}, 2015.

\bibitem{pmlr-v37-rezende15}
Danilo~Jimenez Rezende and Shakir Mohamed.
\newblock {Variational Inference with Normalizing Flows}.
\newblock {\em International Conference on Machine Learning (ICML)}, 2015.

\bibitem{Robinek}
Michal {Rolínek}, Dominik {Zietlow}, and Georg {Martius}.
\newblock {Variational Autoencoders} pursue {PCA} directions (by accident).
\newblock In {\em 2019 IEEE/CVF Conference on Computer Vision and Pattern
  Recognition (CVPR)}, pages 12398--12407, 2019.

\bibitem{salakhutdinov_deep}
Ruslan Salakhutdinov and Geoffrey Hinton.
\newblock {Deep Boltzmann Machines}.
\newblock In {\em {proceedings of the Twelfth International Conference on
  Artificial Intelligence and Statistics}}, volume 5 of JMLR, 2009.

\bibitem{suykensdeep2017}
Johan A.~K. Suykens.
\newblock {Deep Restricted Kernel Machines} using conjugate feature duality.
\newblock {\em Neural Computation}, 29(8):2123--2163, August 2017.

\bibitem{xiao2017/online}
Han Xiao, Kashif Rasul, and Roland Vollgraf.
\newblock {{Fashion-MNIST}: A Novel Image Dataset for Benchmarking Machine
  Learning Algorithms}.
\newblock {\em arXiv:1708.07747}, 2017.

\bibitem{YangPilanci}
Yun Yang, Mert Pilanci, and Martin~J. Wainwright.
\newblock Randomized sketches for kernels: Fast and optimal nonparametric
  regression.
\newblock {\em The Annals of Statistics}, 45(3):991--1023, 2017.

\end{thebibliography}

\section*{Appendix}
\appendix

\section{Proof of Lemma \ref{Lemma:Smoothness}}
We first quote a result that is used in the context of optimization (\cite{Nesterov}, Lemma 1.2.4).
Let $f$ a function with {$L_a$}-Lipschitz continuous Hessian. Then,
\begin{equation}
	\Big|\underbrace{f(\bm{y}_1)-f(\bm{y})-\nabla f(\bm{y})^\top (\bm{y}_1-\bm{y}) -\frac{1}{2}(\bm{y}_1-\bm{y})^\top \Hess_{\bm{y}}[f] (\bm{y}_1-\bm{y})}_{r(\bm{y}_1-\bm{y})}\Big| \leq \frac{{L_a}}{6} \|\bm{y}_1-\bm{y}\|_2^3.\label{eq:Series}
\end{equation}
Then, we calculate the power series expansion of $f(\bm{y}) = [\bm{x}-\bm{\psi}(\bm{y})]_a^2$ and take the expectation with respect to $\bm{\epsilon}\sim \mathcal{N}(0,\mathbb{I})$.
First, we have $
	\nabla f(\bm{y}) =- 2[\bm{x}-\bm{\psi}(\bm{y})]_a\nabla \bm{\psi}_a(\bm{y})$
and
\begin{equation*}
	\Hess_{\bm{y}}[f] = 2\nabla \bm{\psi}_a(\bm{y}) \nabla \bm{\psi}_a(\bm{y})^\top - 2 [\bm{x}-\bm{\psi}(\bm{y})]_a \Hess_{\bm{y}}[\bm{\psi}_a].
\end{equation*}
Then, we use~\eqref{eq:Series} with $\bm{y}_1-\bm{y} = \sigma U\bm{\epsilon}$. By taking the expectation over $\bm{\epsilon}$, notice that the order 1 term in $\sigma$ vanishes since $\mathbb{E}_{\bm{\epsilon}}[\bm{\epsilon}] = 0$.
We find
\begin{align*}
	\mathbb{E}_{\bm{\epsilon}} [\bm{x} - \bm{\psi}(\bm{y}+\sigma  U\bm{\epsilon})]_a^2
	= & [\bm{x} - \bm{\psi}(\bm{y})]_a^2 +\sigma^2  \Tr\big(U^\top\nabla\bm{\psi}_a(\bm{y})\nabla\bm{\psi}_a(\bm{y})^\top U\big)\label{eq:Smoothness}   \\
	- & \sigma^2 [ \bm{x}- \bm{\psi}(\bm{y})]_a \Tr\big(U^\top\Hess_{\bm{y}} [\bm{\psi}_a] U \big)+\mathbb{E}_{\bm{\epsilon}}r(\sigma  U\bm{\epsilon}),
\end{align*}
where we used that $\mathbb{E}_{\bm{\epsilon}}[\bm{\epsilon}^\top M \bm{\epsilon}] = \Tr[M]$ for any symmetric matrix $M$ since $\mathbb{E}_{\bm{\epsilon}}[\bm{\epsilon}_i \bm{\epsilon}_j] = \delta_{ij}$.
Next, denote $R_a(\sigma) = \mathbb{E}_{\bm{\epsilon}}r(\sigma  U\bm{\epsilon})$  we can use the Jensen inequality and subsequently \eqref{eq:Series}
\[|R_a(\sigma)|=
	|\mathbb{E}_{\bm{\epsilon}}r(\sigma  U\bm{\epsilon})|\leq \mathbb{E}_{\bm{\epsilon}}|r(\sigma  U\bm{\epsilon})|\leq \frac{{L_a}}{6} \mathbb{E}_{\bm{\epsilon}}\|\sigma  U\bm{\epsilon}\|_2^3.
\]
Next, we notice that $\|\sigma  U\bm{\epsilon}\|_2 = \sigma (\bm{\epsilon}^\top U^\top U\bm{\epsilon})^{1/2} = \sigma \| \bm{\epsilon}\|_2 $. It is useful to notice that $\| \bm{\epsilon}\|_2$ is distributed according to a chi distribution. By using this remark, we find
\[|R_a(\sigma)|\leq \sigma^3\frac{{L_a}}{6} \mathbb{E}_{\bm{\epsilon}}\|\bm{\epsilon}\|_2^3 = \sigma^3\frac{{L_a}}{6} \frac{\sqrt{2}(m+1) \Gamma((m+1)/2)}{\Gamma(m/2)},
\]
where the last equality uses the expression for the third moment of the chi distribution and where the Gamma function $\Gamma$ is the extension of the factorial to the complex numbers.
\begin{table}[h]
	\caption{Datasets and hyperparameters used for the experiments. $N$  is the number of training samples, $d$ the input dimension (resized images), $m$ the subspace dimension and $M$ the minibatch size.}
	\label{Table:dataset}
	\centering
	\begin{tabular}{lccccc}
		\toprule
		\textbf{Dataset} & $N$    & $d$                     & $m$ & $M$ \\ \midrule
		MNIST            & 60000  & $28 \times 28$          & 10  & 256 \\
		fMNIST           & 60000  & $28 \times 28$          & 10  & 256 \\
		SVHN             & 73257  & $32 \times 32 \times 3$ & 10  & 256 \\
		Dsprites         & 737280 & $64 \times 64$          & 5   & 256 \\
		3Dshapes         & 480000 & $64 \times 64 \times 3$ & 6   & 256 \\
		Cars3D           & 17664  & $64 \times 64 \times 3$ & 3   & 256 \\ \bottomrule
	\end{tabular}
\end{table}

\section{Details on Evidence Lower Bound for \texorpdfstring{$\St$-} RRKM model} \label{sec: details_elbo}

Now we discuss the details of ELBO given in section~\ref{sec:elbo}. The first term in~\eqref{eq:ProbaRKM0} is
\begin{align*}
	\mathbb{E}_{ q_U(\bm{z}|\bm{x}_i)}[\log (p(\bm{x}_i|\bm{z}))] = & - \frac{1}{2\sigma_0^2}\mathbb{E}_{\bm{\epsilon}\sim\mathcal{N}(0,\mathbb{I})}\|\bm{x}_i - \bm{\psi}_{\bm{\xi}}(\mathbb{P}_U\bm{\phi}_{\bm{\theta}}(\bm{x}_i)+\sigma\mathbb{P}_U\bm{\epsilon} + \delta\mathbb{P}_{U^\perp}\bm{\epsilon})\|_2^2 \\
	                                                                & -\frac{d}{2}\log(2\pi\sigma_0^2),
\end{align*}
where we used the following reparameterization following~\cite{kingma_auto_encoding_2013}: $\mathbb{E}_{q_{U}\left(\mathbf{z} \mid \bm{x}_i\right)}[f(\mathbf{z})]=\mathbb{E}_{\bm{\epsilon}\sim\mathcal{N}(0,\mathbb{I})}\left[f\left(\mathbb{P}_U\bm{\phi}_{\bm{\theta}}(\bm{x}) + (\sigma\mathbb{P}_U+\delta \mathbb{P}_{U^\perp})\epsilon\right)\right]$, with $p(\bm{x}|\bm{z}) =  \mathcal{N}(\bm{x}|\bm{\psi_{\xi}}(\bm{z}),\sigma^2_0\mathbb{I})$ and $q_U(\bm{z}|\bm{x})=\mathcal{N}(\bm{z}|\mathbb{P}_U\bm{\phi}_{\bm{\theta}}(\bm{x}),\sigma^2\mathbb{P}_U+\delta^2 \mathbb{P}_{U^\perp})$.
Clearly, the above expectation can be written as follows
\begin{equation*}
	\mathbb{E}_{\bm{\epsilon}}\mathbb{E}_{\bm{\epsilon}_\perp}\|\bm{x}_i - \bm{\psi}_{\bm{\xi}}(\mathbb{P}_U\bm{\phi}_{\bm{\theta}}(\bm{x}_i)+
	\sigma U\bm{\epsilon} + \delta U_\perp\bm{\epsilon}_\perp)\|_2^2,
\end{equation*}
with $\bm{\epsilon}\sim\mathcal{N}(0,\mathbb{I}_m)$ and $\bm{\epsilon}_\perp \sim\mathcal{N}(0,\mathbb{I}_{\ell -m})$.
Hence, we fix $\sigma_0^2 = 1/2$ and take $\delta>0$ to a numerically small value.
For the other terms of~\eqref{eq:ProbaRKM0}, we use the formula giving the KL divergence between multivariate normals. Let $\mathcal{N}_0$ and $\mathcal{N}_1$ be $\ell$-variate normal distributions with mean $\mu_0, \mu_1$ and covariance $\Sigma_0, \Sigma_1$ respectively.
Then,
\begin{equation*}
	\KL(\mathcal{N}_0,\mathcal{N}_1) = \frac{1}{2}\Big\{\tr(\Sigma_1^{-1}\Sigma_0) \\+ (\mu_1-\mu_0)^\top \Sigma_1^{-1}(\mu_1-\mu_0)-\ell + \log\left(\frac{\det\Sigma_1}{\det\Sigma_0}\right) \Big\}
\end{equation*}
By using this identity, we find  the second term of~\eqref{eq:ProbaRKM0}:
\begin{align*}
	\KL[q_U(\bm{z}|\bm{x}_i), q(\bm{z}|\bm{x}_i)] = \frac{1}{2}\Big\{ & \frac{m\sigma^2 + (\ell-m) \delta^2}{\gamma^2}+
	\frac{1}{\gamma^2}\|\bm{\phi}_{\bm{\theta}}(\bm{x}_i)-\mathbb{P}_U \bm{\phi}_{\bm{\theta}}(\bm{x}_i)\|_2^2                                             \\
	                                                                  & -\ell +\log\big( \frac{\gamma^{2\ell}}{\sigma^{2m}\delta^{2(\ell-m)}}\big) \Big\},
\end{align*}
{where $q(\bm{z}|\bm{x})=\mathcal{N}(\bm{z}|\bm{\phi}_{\bm{\theta}}(\bm{x}),\gamma^2\mathbb{I}_\ell)$.}
For the third term in~\eqref{eq:ProbaRKM0}, we find
\begin{align*}
	\KL[q_U(\bm{z}|\bm{x}_i), p(\bm{z})] = \frac{1}{2}\Big\{\Tr((\sigma^{2}\mathbb{P}_U + \delta^2\mathbb{P}_{U^\perp})\Sigma^{-1}) +
	(\mathbb{P}_U \bm{\phi}_{\bm{\theta}}(\bm{x}_i))^\top \Sigma^{-1}(\mathbb{P}_U \bm{\phi}_{\bm{\theta}}(\bm{x}_i)) \\
	+\log\det(\Sigma)  -\ell -\log (\sigma^{2m}\delta^{2(\ell-m)})\Big\},
\end{align*}
{with $p(\bm{z}) = \mathcal{N}(0,\Sigma)$.}
By averaging over $i=1,\dots,n$, we obtain
\begin{align*}
	\frac{1}{n}\sum_{i=1}^n \KL[q_U(\bm{z}|\bm{x}_i), p(\bm{z})] = \frac{1}{2}\Big\{ \Tr((\sigma^{2}\mathbb{P}_U + \delta^2\mathbb{P}_{U^\perp})\Sigma^{-1})
	+ \Tr(\mathbb{P}_{U}C_{\bm{\theta}}\mathbb{P}_{U} \Sigma^{-1}) \\ +\log\det(\Sigma)
	-\ell -\log(\sigma^{2m}\delta^{2(\ell-m)})\Big\},
\end{align*}
where we used the cyclic property of the trace and $C_{\bm{\theta}}= \frac{1}{n} \sum_{i=1}^n\bm{\phi}_{\bm{\theta}}(\bm{x}_i)\bm{\phi}_{\bm{\theta}}(\bm{x}_i)^\top$.
This proves the analogous expression in section~\ref{sec:elbo}. Finally, the estimation of the optimal $\Sigma$ can be done in parallel to the Maximum Likelihood Estimation of the covariance matrix of a multivariate normal.
\begin{table}[ht]
	\caption{Model architectures. All convolutions and transposed-convolutions are with stride 2 and padding 1. Unless stated otherwise, layers have Parametric-RELU ($\alpha = 0.2$) activation functions, except output layers of the pre-image maps which have sigmoid activation functions (since input data is normalized $[0,1]$). Adam and CayleyAdam optimizers have learning rates $2\times 10^{-4}$ and $10^{-4}$ respectively. Pre-image map/decoder network is always taken as transposed of feature map/encoder network. $c=48$ for Cars3D; and $c=64$ for all others. Further, $\hat{k}=3$ and stride 1 for MNIST, fMNIST, SVHN and 3Dshapes; and $\hat{k}=4$ for others. SVHN and 3Dshapes are resized to $28\times 28 $ input dimensions. \label{tab:arch}}
	\centering
	\resizebox{\textwidth}{!}{
		\begin{tabular}{l l l}
			\toprule
			\textbf{Dataset} & \multicolumn{2}{c}{\textbf{Architecture}} \\
			\midrule
			\makecell[l]{MNIST/fMNIST/                                   \\/SVHN/3Dshapes/\\Dsprites/Cars3D}         & $\bm \phi_{\theta}(\cdot) = \begin{cases} \makecell[l]{Conv~ [c]\times 4 \times 4; \\Conv ~[c \times 2] \times 4 \times 4;\\ Conv~ [c \times 4] \times \hat{k}\times\hat{k};     \\  FC~ 256; \\ FC~ 50 ~(Linear)}    \end{cases} $ & $ \bm \psi_{\zeta}(\cdot) = \begin{cases}
					FC~256;                                       \\
					FC~ [c \times 4] \times \hat{k}\times\hat{k}; \\
					Conv ~[c \times 2] \times 4 \times 4;         \\
					Conv ~[c] \times 4 \times 4;                  \\
					Conv ~[c]~ (Sigmoid)                          \\
				\end{cases} $ \\
			\bottomrule
		\end{tabular}
	}
\end{table}
\section{Datasets and Hyperparameters}
\label{sec:hyperparams}
We refer to Table~\ref{Table:dataset}  and  Table~\ref{tab:arch}  for specific  details  on  model  architectures, datasets and hyperparameters used in this paper. All models were trained on full-datasets and for maximum 1000 epochs. Further all datasets are scaled between [0-1] and are resized to $28\times 28$ dimensions except Dsprites and Cars3D. The PyTorch library (single precision) in Python was used as the programming language on 8GB NVIDIA QUADRO P4000 GPU. See Algorithm~\ref{algo} for training the $\St$-RKM model. In the case of FactorVAE, the discriminator architecture is same as proposed in the original paper~\cite{FactorVAE}.

\textbf{Disentanglement Metrics:}
MIG was originally proposed by~\cite{MIG_VAE}, however we use the modified metric as proposed in \cite{locatelloChallengingassumptions}. Further we evaluate this score on 5000 test points across all the considered datasets.
SAP and Eastwood's metrics both uses different classifiers to compute the importance of each dimension of the learned representation for predicting a ground-truth factor. For these metrics, we randomly sample 5000 and 3000 training and testing points respectively. To compute these metrics, we use the open source library available at github.com/google-research/disentanglement\_lib.

\section{Ablation studies}
\label{sec:ablation_study}
\noindent \textbf{Significance of the KPCA loss:}
In this section, we show an ablation study on the KPCA loss and evaluate its effect on disentanglement. We repeat the experiments of Section~\ref{sec:exp} on the \emph{mini}-3DShapes dataset (floor hue, wall hue, object hue and scale: 8000 samples), where we consider 3 different variants of the proposed model:
\begin{enumerate}
	\item \textbf{St-RKM} ($\sigma = 0$): The KPCA loss is optimized in a stochastic manner using the Cayley Adam optimizer, as proposed in the paper.
	\item \textbf{Gen-RKM:} The KPCA loss is optimized exactly at each step by performing an eigendecomposition in each mini-batch (this corresponds to the algorithm in~\cite{GENRKM}).
	\item \textbf{AE-PCA:} A standard AE is used and a reconstruction loss is minimized for the training. As a post-processing step, a PCA is performed on the latent embedding of the training data.
\end{enumerate}
The encoder/decoder maps are the same across all the models, and for the AE-PCA model, additional linear layers are used to map the latent space to the subspace.
From Table~\ref{tab:rkm_ablation}, we conclude that optimizing the KPCA loss during training improves disentanglement. Moreover, using a stochastic algorithm improves computation time and scalability with only a slight decrease in disentanglement score. Note that calculating the exact eigendecomposition at each step (Gen-RKM) comes with numerical difficulties. In particular, double floating-point precision has to be used together with a careful selection of the number of principal components to avoid ill-conditioned kernel matrices. This problem is not encountered when using the St-RKM training algorithm.
\begin{table}[ht]
	\centering
	\caption{Training timings per epoch (in minutes) and disentanglement scores (\cite{Heusel2017}) for different variants of RKM when trained on the \emph{mini}-3Dshapes dataset. Gen-RKM has the worst training time but gets the highest disentanglement scores. This is due to the exact eigendecomposition of the kernel matrix at every iteration. This computationally expensive step is approximated by the St-RKM model which achieves significant speed-up and scalability to large datasets. Finally, the AE-PCA model has the fastest training time due to the absence of eigendecompositions in the training loop. However, using PCA in the post-processing step alters the basis of the latent-space. This basis is unknown to the decoder network resulting in degraded disentanglement performance.}
	\begin{tabular}{llccc}
		\toprule
		                               &       & $\St$\textbf{-RKM} ($\sigma = 0$) & \textbf{Gen-RKM}     & \textbf{AE-PCA}      \\ \midrule \midrule
		Train. time                    &       & 3.01~(0.71)                       & 9.21~(0.54)          & \textbf{2.87}~(0.33) \\\midrule \midrule
		%   FID    & & & \\
		\multirow{2}{*}{Disent. score} & Lasso & 0.40~(0.02)                       & \textbf{0.44}~(0.01) & 0.35~(0.01)          \\ %\cmidrule{2-5}
		                               & RF    & 0.27~(0.01)                       & \textbf{0.31}~(0.02) & 0.22~(0.02)          \\ \midrule
		\multirow{2}{*}{Compl. score}  & Lasso & \textbf{0.64}~(0.01)              & 0.51~(0.01)          & 0.42~(0.01)          \\
		                               & RF    & \textbf{0.67}~(0.02)              & 0.58~(0.01)          & 0.45~(0.02)          \\\midrule
		\multirow{2}{*}{Info. score}   & Lasso & \textbf{1.01}~(0.02)              & 1.11~(0.02)          & 1.20~(0.01)          \\
		                               & RF    & \textbf{0.98}~(0.01)              & 1.09~(0.01)          & 1.17~(0.02)          \\ \bottomrule
	\end{tabular}
	\label{tab:rkm_ablation}
\end{table}
%\alert{awd}
\begin{table}[h]
	\centering
	\caption{FID scores (lower is better, with standard deviations) computed on randomly generated 8000 images when trained with architecture and hyperparameters adapted from \cite{dupont2018learning}.}
	% \resizebox{\textwidth}{!}{%
		\begin{tabular}{lccccc}
			\toprule
			       & St-\textbf{RKM}       & \textbf{VAE} & $\beta$-\textbf{VAE} & \textbf{FactorVAE} & \textbf{InfoGAN} \\ \midrule
			MNIST  & \textbf{24.63}~(0.22) & 36.11~(1.01) & 42.81~(2.01)         & 35.48~(0.07)       & 45.74~(2.93)     \\
			fMNIST & \textbf{61.44}~(1.02) & 73.47~(0.73) & 75.21~(1.11)         & 69.73~(1.54)       & 84.11~(2.58)     \\ \bottomrule
		\end{tabular}
	% }
%
	\label{tab:smaller_arch}
\end{table}

\begin{table*}[ht]
	\centering
	\caption{Computing the diagonalization scores (see Figure~\ref{fig:resultant_matrices}).
	Denote $M = \frac{1}{|\mathcal{C}|}\sum_{i\in \mathcal{C}}U^{\top}_{\star}\nabla\bm{\psi}(\bm{y}_i)\nabla\bm{\psi}(\bm{y}_i)^\top U_\star, \text{ with } \bm{y}_i = \mathbb{P}_{U}\bm{\phi_{\theta}}(\bm{x}_i)$ (cf.~\eqref{eq:MC_covariance_of_variations}).
	Then we compute the score as $ \left\| M - \mathrm{diag}(M)  \right\|_{{{F}}} / \left\|  M \right\|_{{F}}$, where $ \mathrm{diag} \colon \mathbb{R}^{m\times m} \mapsto \mathbb{R}^{m\times m} $ sets the off-diagonal elements of matrix to zero. The scores are computed for each model over 10 random seeds and show the mean (standard deviation).
	Lower scores indicate better diagonalization.}
	\label{Table:diag_scores}
	\begin{tabular}{lccc}
		\toprule
		\textbf{Models}                                     & \multicolumn{1}{c}{\textbf{Dsprites}} & \multicolumn{1}{c}{\textbf{3Dshapes}} & \multicolumn{1}{c}{\textbf{Cars3D}} \\ \midrule
		{$\St$-RKM-sl} ($\sigma = 10^{-3}$, $ U_{\star}  $) & \textbf{0.17} (0.05)                  & \textbf{0.23} (0.03)                  & \textbf{0.21} (0.04)                \\
		{$\St$-RKM} ($\sigma = 10^{-3}$, $ U_{\star}  $)    & 0.26 (0.05)                           & 0.30 (0.10)                           & 0.31 (0.09)                         \\
		{$\St$-RKM} ($\sigma=10^{-3}$, random $U$)          & 0.61 (0.02)                           & 0.72 (0.01)                           & 0.69 (0.03)
	\end{tabular}
\end{table*}

\begin{figure}[h]
	\centering
	\setlength{\tabcolsep}{1pt}
	\resizebox{\textwidth}{!}{
		\begin{tabular}{r c c c c}
			 & $\St$-RKM ($\sigma=0$)                                    & $\beta$-VAE ($\beta=3$)                          & Info-GAN & FactorVAE \\
			\rotatebox[origin=c]{90}{MNIST}
			 & \tabfigure{width=5cm}{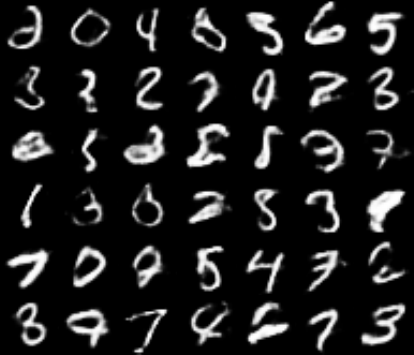}
			 & \tabfigure{width=5cm}{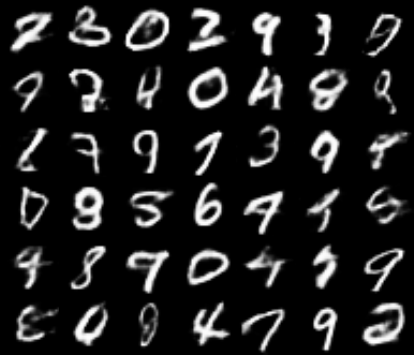}    & \tabfigure{width=5cm}{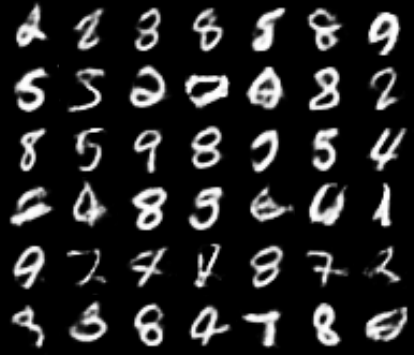}
			 & \tabfigure{width=5cm}{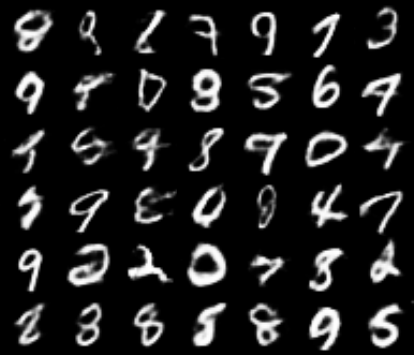}                                                                                  \\
			 &                                                           &                                                  &          &           \\
			\rotatebox[origin=c]{90}{fMNIST}
			 & \tabfigure{width=5cm}{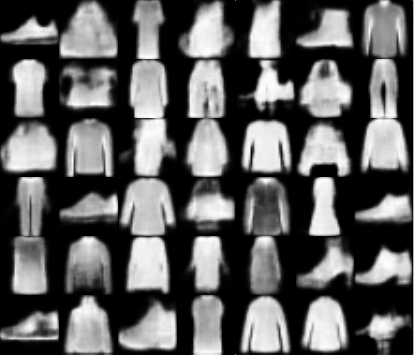}
			 & \tabfigure{width=5cm}{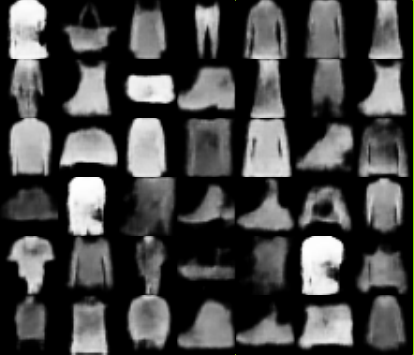}
			 & \tabfigure{width=5cm}{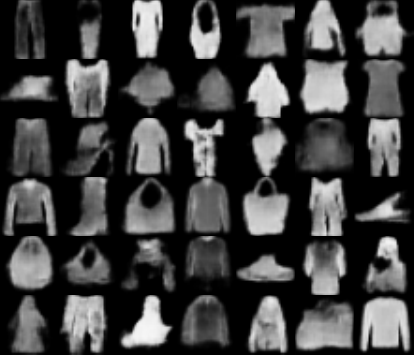}
			 & \tabfigure{width=5cm}{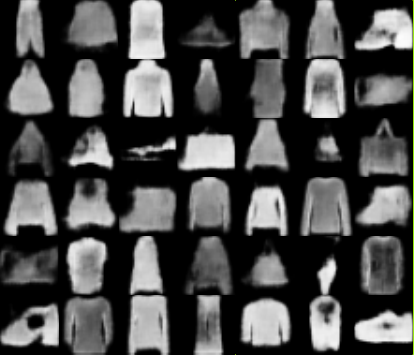}
		\end{tabular}
	}
	\caption{Samples of randomly generated %batch of
		images used to compute the FID scores; see Table~\ref{tab:smaller_arch}.}
	\label{fig:smaller_arch}
\end{figure}

\noindent\textbf{Smaller encoder/decoder architecture:}
	In this section, we analyze the impact of the encoder/decoder architecture on the generation quality of considered models. The generation quality experiment of Section~\ref{sec:exp} is repeated on the fMNIST and MNIST dataset, where the architecture and hyperparameters are adapted from~\cite{dupont2018learning}. From Table~\ref{tab:smaller_arch} and Figure~\ref{fig:smaller_arch}, we see that the overall FID scores and generation quality have improved, however, the relative scores among the models did not change significantly. 

\noindent\textbf{Analysis of St-RKM with a fixed $U$ :}
We discuss here the role of the optimization of $\St(\ell, m)$ on disentanglement in the case of a classical AE loss ($\sigma = 0$). To do so, a matrix $\tU\in\St(\ell, m)$ is generated randomly\footnote{Using a random $\tU\in\St(\ell, m)$ can be interpreted as sketching the encoder map in the spirit of Randomized Orthogonal Systems (ROS) sketches (see~\cite{YangPilanci}).} and kept fixed during the training of the following optimization problem
\begin{equation}
	\min_{\substack{\bm{\theta}, \bm{\xi}}}\lambda\frac{1}{n}\sum_{i=1}^{n} L^{(0)}_{\bm{\xi},\tU}\left(\bm{x}_i,\bm{\phi}_{\bm{\theta}}(\bm{x}_i)\right) + \underbrace{\frac{1}{n}\sum_{i=1}^{n}\|\mathbb{P}^{(\varepsilon)}_{\tU^\perp} \bm{\phi}_{\bm{\theta}}(\bm{x}_i)\|_2^2}_{\text{regularized PCA objective}},\label{eq:ReducedObjectiveFrozen}
\end{equation}
with $\lambda = 1$ and where $\varepsilon\geq 0$ is a regularization constant and where the  regularized (or mollified) projector $ \mathbb{P}_{\tU^\perp}^{(\varepsilon)} = \varepsilon ( \tU\tU^\top +\varepsilon \mathbb{I}_{\ell})^{-1} $
is used in order to prevent numerical instabilities. Indeed, if $\varepsilon=0$, the second term in~\eqref{eq:ReducedObjectiveFrozen} (PCA term)  is not strictly convex as  a function of $\bm{\phi}_{\bm{\theta}}$, since  this quadratic form has flat directions along the column subspace of $\tU$. Our numerical simulations in single-precision PyTorch with $\varepsilon=0$ exhibit instabilities, i.e.,  the PCA term in~\eqref{eq:ReducedObjectiveFrozen} takes negative values during the training. 
Hence, the regularized projector is introduced so that the PCA quadratic is strongly convex for $\varepsilon > 0$. This instability is not observed in the training of~\eqref{eq:ReducedObjective} where $U$ is not fixed.
This is one asset of our training procedure using optimization over Stiefel manifold. Explicitly, the regularized projector satisfies the following properties
\begin{itemize}
	\item $\mathbb{P}_{\tU^\perp}^{(\varepsilon)} u_\perp = u_\perp$ for all $u_\perp \in (\range(U))^\perp$,
	\item $\mathbb{P}_{\tU^\perp}^{(\varepsilon)} u = \varepsilon u$ for all $u \in \range(U)
	      $.
\end{itemize}
Thanks to the push-through identity,  we have the alternative expression $
	\mathbb{P}_{\tU^\perp}^{(\varepsilon)}  = \mathbb{I} - U (U^\top U + \varepsilon \mathbb{I}_{m})^{-1} U^\top.$ Therefore, it holds $\lim_{\varepsilon\to 0} \mathbb{P}_{\tU^\perp}^{(\varepsilon)} = \mathbb{P}_{\tU^\perp}$, as it should.
In our experiments, we set $\varepsilon = 10^{-5}$. If $\varepsilon \leq 10^{-6}$, the regularized PCA objective in~\eqref{eq:ReducedObjectiveFrozen} takes negative values after a few epochs due to the numerical instability as mentioned above.

\begin{figure}[h]
	\begin{subfigure}[b]{0.48\textwidth}
		\centering
    \includegraphics[width=\textwidth]{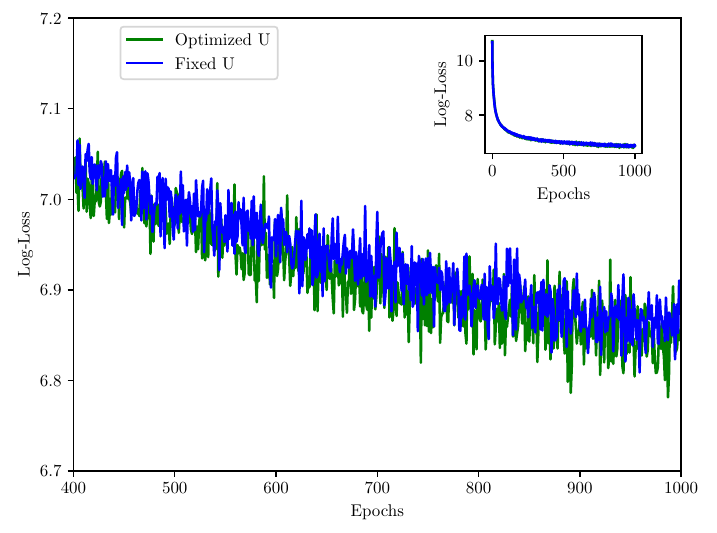}
		\caption{Loss-evolution}
		\label{fig:Loss_U_FroU}
	\end{subfigure}
	\begin{subfigure}[b]{0.48\textwidth}
		\centering
		\tabfigure{trim={0 -0.6cm 0 0cm}, clip, width=6.5cm}{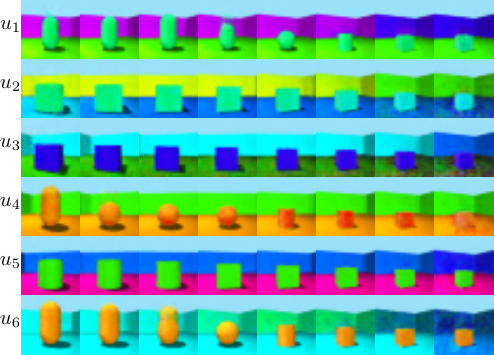}
		\caption{Latent space Traversal}
		\label{fig:traversal_U_FroU}
	\end{subfigure}
	\caption{(a) Loss evolution ($\log$ plot) during the training of \eqref{eq:ReducedObjectiveFrozen} over 1000 epochs with $\varepsilon = 10^{-5}$ once with Cayley-Adam optimizer (green curve) and then without (blue curve).  (b) Traversals along the principal components when the model was trained with a fixed $U$, i.e. with the objective given by \eqref{eq:ReducedObjectiveFrozen} and $\varepsilon = 10^{-5}$. There is no clear isolation of a feature along any of the principal components indicating further that optimizing over $U$ is key for better disentanglement.}
	\label{fig:fig}
\end{figure}

\begin{figure}[ht]
	\centering
	\setlength{\tabcolsep}{1pt}
	\resizebox{\textwidth}{!}{
		\begin{tabular}{r c c c c}
			                                 & $\St$-RKM ($\sigma=0$)                                      & $\beta$-VAE ($\beta=3$)                                  & Info-GAN                                             & FactorVAE \\
			\rotatebox[origin=c]{90}{MNIST}  & \tabfigure{width=5cm}{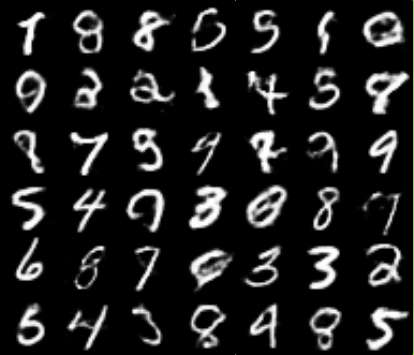}        & \tabfigure{width=5cm}{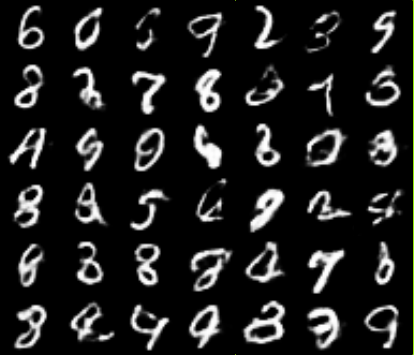} & \tabfigure{width=5cm}{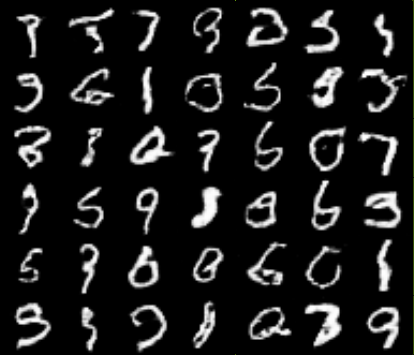}
			                                 & \tabfigure{width=5cm}{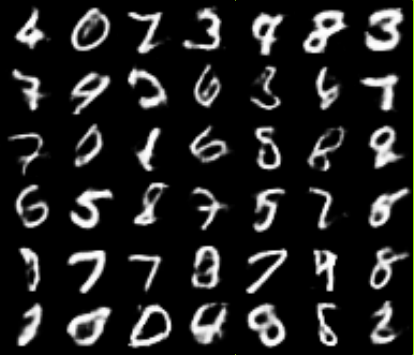}                                                                                                                                   \\
			                                 &                                                             &                                                          &                                                      &           \\
			\rotatebox[origin=c]{90}{fMNIST} & \tabfigure{width=5cm}{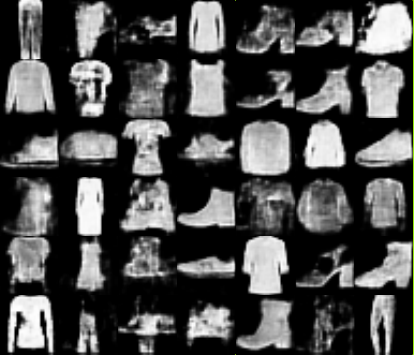}         & \tabfigure{width=5cm}{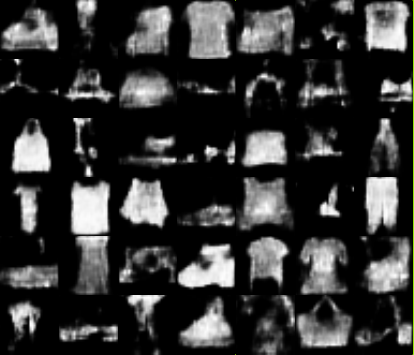} & \tabfigure{width=5cm}{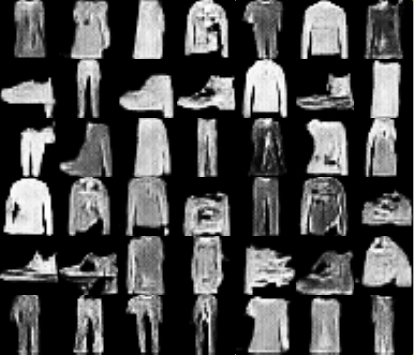}
			                                 & \tabfigure{width=5cm}{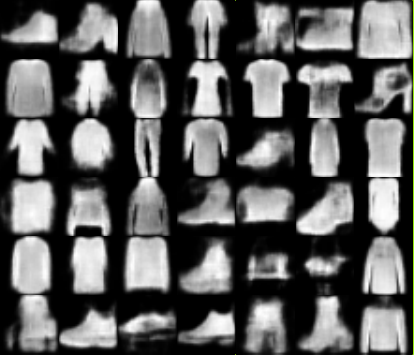}                                                                                                                                 \\
			                                 &                                                             &                                                          &                                                      &           \\
			\rotatebox[origin=c]{90}{SVHN}
			                                 & \tabfigure{width=5cm}{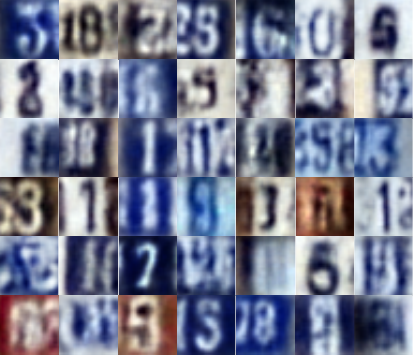}
			                                 & \tabfigure{width=5cm}{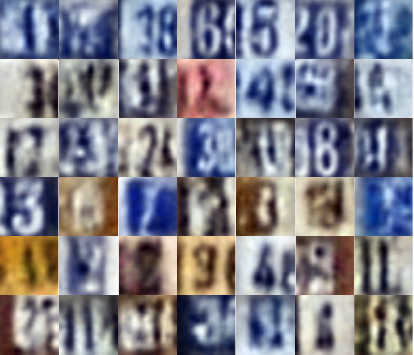}       & \tabfigure{width=5cm}{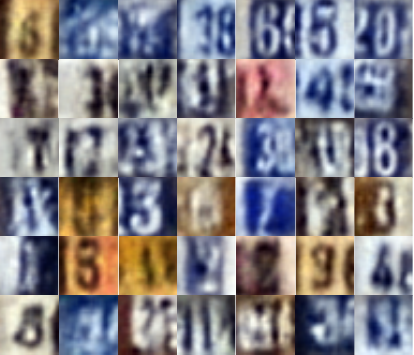}
			                                 & \tabfigure{width=5cm}{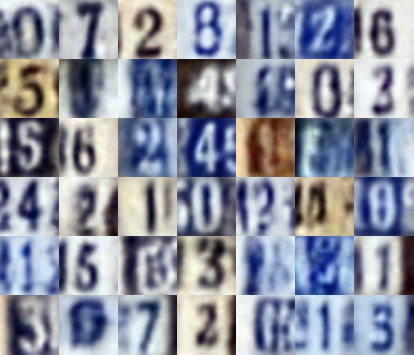}                                                                                                                                  \\
			                                 &                                                             &                                                          &                                                      &           \\
			\rotatebox[origin=c]{90}{3Dshapes}
			                                 & \tabfigure{width=5cm}{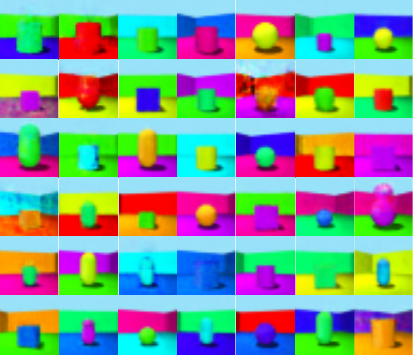}
			                                 & \tabfigure{width=5cm}{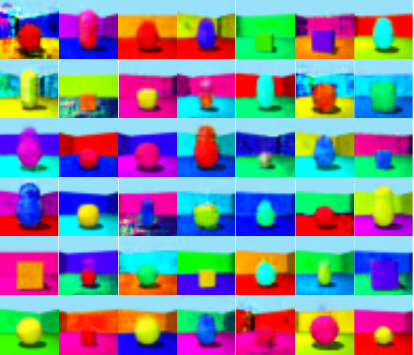}       & \tabfigure{width=5cm}{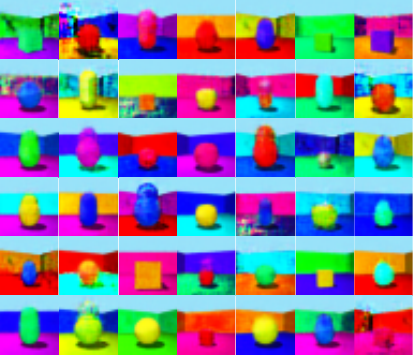}
			                                 & \tabfigure{width=5cm}{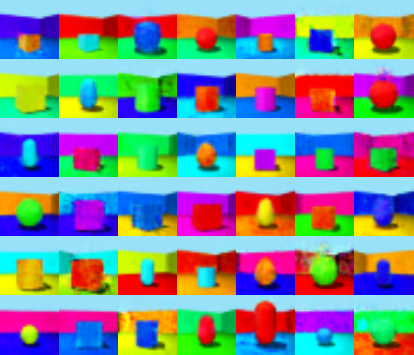}                                                                                                                               \\
			                                 &                                                             &                                                          &                                                      &           \\
			\rotatebox[origin=c]{90}{Dsprites}
			                                 & \tabfigure{width=5cm}{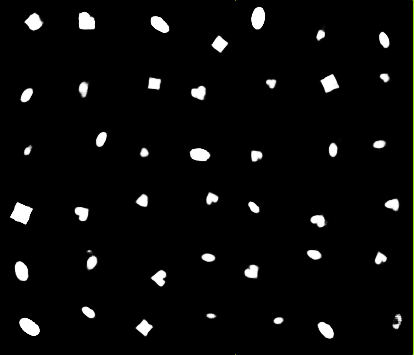}
			                                 & \tabfigure{width=5cm}{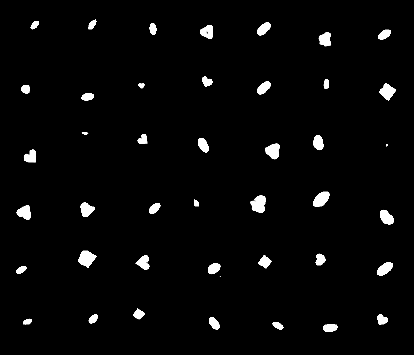}  & \tabfigure{width=5cm}{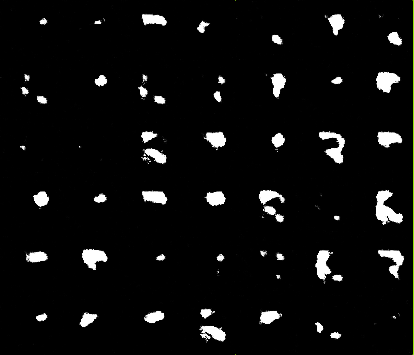}
			                                 & \tabfigure{width=5cm}{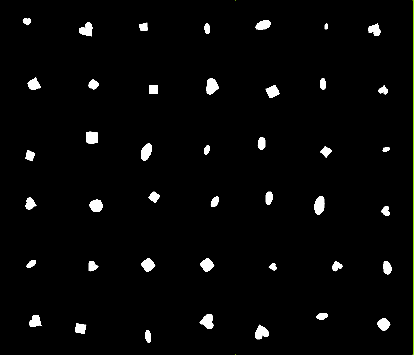}
		\end{tabular}
	}
	\caption{Samples of randomly generated batch of images used to compute FID scores and SWD scores (see Fig. ~\ref{fig:fid_swd}).}
	\label{fig:fid_sample_imgs}
\end{figure}
In Figure~\ref{fig:Loss_U_FroU}, the evolution of the training objective~\eqref{eq:ReducedObjectiveFrozen} is displayed.  It can be seen that the final objective has a lower value [$\exp(6.78) \approx 881$] when $U$ is optimized compared to its fixed counterpart [$\exp(6.81) \approx 905$] showing the merit of  optimizing over Stiefel manifold for the same parameter $\varepsilon$. Hence the subspace determined by $\range(U)$ has to be adapted to the encoder and decoder networks. In other words, the training over $\bm{\theta}, \bm{\xi}$ is not sufficient to minimize the $\St(\ell, m)$ objective with Adam. Figure~\ref{fig:traversal_U_FroU} further explores the latent traversals in the context of this ablation study. In the top row of Figure~\ref{fig:traversal_U_FroU} (latent traversal in the direction of $\bm{u}_1$), both the shape of the object and the wall hue are changing. A coupling between wall hue and shape is also visible in the bottom row of this figure.

%%%%%%%

\end{document}